\def\year{2020}\relax
\documentclass[letterpaper]{article} % DO NOT CHANGE THIS
\usepackage{aaai20}  % DO NOT CHANGE THIS
\usepackage{times}  % DO NOT CHANGE THIS
\usepackage{helvet} % DO NOT CHANGE THIS
\usepackage{courier}  % DO NOT CHANGE THIS
\usepackage[hyphens]{url}  % DO NOT CHANGE THIS
\usepackage{graphicx} % DO NOT CHANGE THIS
\usepackage{amsmath,amssymb}
\usepackage{amsthm}
\usepackage{amsfonts}       % blackboard math symbols
\usepackage{mathtools}
\usepackage{nicefrac}
\usepackage{algorithm,algorithmic}
\usepackage{pgfplots, pgfplotstable}

\newcommand{\R}{\mathbb{R}}

\newcommand{\1}{{\mathbf 1}}

\newcommand{\bfT}{\mathbf{T}}

\newcommand{\bfp}{\mathbf{p}}
\newcommand{\bfq}{\mathbf{q}}
\newcommand{\bfx}{\mathbf{x}}
\newcommand{\bfX}{\mathbf{X}}
\newcommand{\bfy}{\mathbf{y}}
\newcommand{\bfY}{\mathbf{Y}}
\newcommand{\bfz}{\mathbf{z}}
\newcommand{\bfr}{\mathbf{r}}

\newcommand{\bfu}{\mathbf{u}}

\newcommand{\bfU}{\mathbf{U}}

\newcommand{\calX}{\mathcal{X}}
\newcommand{\calU}{\mathcal{U}}

\DeclareMathOperator*{\Exp}{\mathbb E}

\DeclareMathOperator*{\argmax}{argmax}

\DeclareMathOperator*{\eu}{\mathit{EU}}
\DeclareMathOperator*{\eus}{\mathit{EU^*}}
\DeclareMathOperator*{\peu}{\mathit{PEU}}
\DeclareMathOperator*{\evoi}{\mathit{EVOI}}
\newcommand{\qs}{\mathcal{Q}} % query space
\newcommand{\rs}{\mathcal{R}} % response space

\newcommand{\Regret}{\mathit{Regret}}

\newcommand{\DR}{\mathsf{DeepRetr}}
\newcommand{\freeevoi}{\mathsf{ContFree}}
\newcommand{\alterevoi}{\mathsf{ContAlter}}
\newcommand{\regevoi}{\mathsf{ContReg}}
\newcommand{\drevoi}{\mathsf{ContDeepRetr}}
\newcommand{\druniq}{\mathsf{DeepRetrUniq}}
\newcommand{\paolo}{\mathsf{RandUserTopItem}}
\newcommand{\contpartial}{\mathsf{ContPartial}}
\newcommand{\greedy}{\mathsf{Greedy}}
\newcommand{\random}{\mathsf{Random}}
\newcommand{\qi}{\mathsf{QueryIteration}}
\newcommand{\topfive}{\mathsf{Top5Exhaustive}}
\newcommand{\balanced}{\mathsf{Balanced}}
\newcommand{\opt}{\mathsf{ExhaustiveSearch}}

\newtheorem{thm}{Theorem}
\newtheorem{definition}[thm]{Definition}

\newtheoremstyle{TheoremNum}%
    {\topsep}{\topsep}%%% space between body and thm
    {\itshape}%%% Thm body font
    {}%%% Indent amount (empty = no indent)
    {\bfseries}%%% Thm head font
    {.}%%% Punctuation after thm head
    { }%%% Space after thm head
    {\thmname{#1}\thmnote{ \bfseries #3}}%%% Thm head spec
\theoremstyle{TheoremNum}

\usepgfplotslibrary{fillbetween}
\pgfplotsset{compat=1.10}
\newcommand{\errorband}[7]{
\pgfplotstableread{#1}\datatable
   \addplot [name path=pluserror,draw=none,no markers,forget plot] table [x={#2},y expr=\thisrow{#5}] {\datatable};

   \addplot [name path=minuserror,draw=none,no markers,forget plot]
     table [x={#2},y expr=\thisrow{#4}] {\datatable};

   \addplot [forget plot,color=#6,opacity=0.2]
     fill between[on layer={},of=pluserror and minuserror];

   \addplot [#6,mark=#7]
     table [x={#2},y={#3}] {\datatable};
}

\usepackage{filecontents}

\title{Gradient-based Optimization for Bayesian Preference Elicitation}

\author{Ivan Vendrov, Tyler Lu, Qingqing Huang, Craig Boutilier\\
Google Research, Mountain View, California\\
\{\texttt{ivendrov,tylerlu,qqhuang,cboutilier}\}\texttt{@google.com}
} 
\begin{document}

\maketitle

\begin{abstract}

% Preference elicitation is an important aspect of interactive and
    % conversational recommendation systems. With the increasing popularity of
    % personal assistants and voice-based user interfaces, it is critical to
    % design systems that ask only the relevant preference questions to arrive
    % at good recommendations. Elicitation with prior populational preference
    % models allows the system to ask minimal number of questions to
    % sufficiently hone in on a user's type for purposes of recommendation. A
    % common objective for selecting queries is the expected value of
    % information (EVOI), which is computationally prohibitive, especially with
    % a large item space. Our work aims to make Bayesian elicitation more
    % practical in two ways: (1) a novel Monte Carlo continuous EVOI
    % optimization method which is more scalable for large item spaces and (2)
    % proposing and studying partial comparison queries, which are more
    % practical for items with large number of attributes. We show how
    % continuous EVOI can be readily adapted for partial comparisons, and how
    % it can be easily implemented as a differentiable network in modern ML
    % frameworks (e.g. TensorFlow, PyTorch, etc.). Experimental results show
    % that gradient-based EVOI methods can achieve state of the art on some
    % full comparison domains and close to the theoretical optimum performance
    % in the partial comparison case.

Effective techniques for eliciting user preferences have taken on added
importance as recommender systems (RSs) become increasingly interactive and
conversational. A common and conceptually appealing
Bayesian criterion for selecting queries
% to elicit preferences from users 
is \emph{expected value of information (EVOI)}.
Unfortunately, it is computationally prohibitive to construct queries with
maximum EVOI in RSs with large item spaces. We tackle this issue by
introducing a \emph{continuous formulation of EVOI as a differentiable
network} that can be optimized using gradient methods available in modern
machine learning (ML) computational frameworks (e.g., TensorFlow, PyTorch).  We exploit this
to develop a novel, scalable Monte Carlo method for EVOI optimization,
which is more scalable for large item spaces than methods requiring
explicit enumeration of items. While we emphasize the use of this approach
for pairwise (or $k$-wise) comparisons of items, we also demonstrate how
our method can be adapted to queries involving subsets of item attributes
or ``partial items,'' which are often more cognitively manageable for
users.
%when items are characterized by a large number of attributes.
% user-friendly \emph{partial comparison queries}.
% (over subsets of item attributes), which are more cognitively manageable in domains with large numbers of attributes.
Experiments show that our gradient-based EVOI technique achieves
state-of-the-art performance across several domains while scaling
to large item spaces.

\end{abstract}

\section{Introduction}

Rapid advances in AI, machine learning, speech and language technologies have
enabled recommender systems (RSs) to become more conversational and
interactive. Increasingly, users engage RSs using language-based (speech or
text) and multi-modal interfaces
% with them using small devices, voice, and other modalities that limit (or even obviate) communication bandwidth using graphical interfaces.
that have the potential to increase communication bandwidth with users
compared to
passive systems that make recommendations based only on user-initiated
engagement.  In such contexts, actively \emph{eliciting} a user's preferences
with a limited amount of interaction can be critical to the user experience.

\emph{Preference elicitation} has been widely studied in decision analysis and
AI \cite{keeney-raiffa,hamalainen:smc01,urszula:aaai00,preference:aaai02}, but
has received somewhat less attention in the recommender community (we discuss
exceptions below). Bayesian methods have proven to be conceptually appealing
for elicitation; in particular, \emph{expected value  of information (EVOI)}
offers a principled criterion for selecting preference queries and determining
when to make recommendations.
EVOI is relatively practical in domains with small decision spaces and,
as such, can be applied to certain types of content-based RSs,
e.g., those with small 
% databases or 
``catalogs'' of attribute-based items.
% \cite{XXX}. 
However, maximizing EVOI is generally computationally intractable
and often unable to scale to realistic settings with millions of items.

%, as often tackled by commercial RSs.
% For example, recommenders that
% use collaborative-filtering (CF) type models (e.g., standard
% matrix-factorization approaches \cite{SMBPMF} or neural CF \cite{yiEtAl:recsys19}) typically
% embed large numbers of users and items into some high-dimensional latent space,
% making any sort of preference elicitation challenging, EVOI notwithstanding.

% , smartphones and voice assistants where a limited amount of information can be presented at once. Actively eliciting user preferences by asking questions is an increasingly critical mode of interaction vis-a-vis passively waiting for user input.

%  To facilitate good decision-making in a world of increasing information overload, it's critical to be able to ask the right questions-the ones that elicit the most decision-relevant information and that are easiest for the user to answer. The Bayesian criterion of maximizing expected value of information (EVOI) gives a principled approach to finding such questions, but has been hard to scale to realistic settings with large number of items and attributes.
 
Our work aims to make Bayesian elicitation more practical. Our first
contribution is a novel formulation of the problem of selecting a \emph{slate
query with maximum EVOI} as a continuous optimization in a differentiable
network. This allows the problem to be solved using gradient-based techniques
available in modern ML computational frameworks such as TensorFlow or PyTorch.
The key to our approach is relaxing the set of items from which recommendations
and queries are generated by allowing attributes to vary continuously without
requiring they correspond to available items (these could be interpreted
as ``hypothetical'' items in attribute- or embedding-space). Once optimized in
this relaxed fashion, any hypotheticals are projected back into the nearest
actual items to generate suitable recommendations or queries. We also
propose regularizers that can be used during optimization to keep hypothetical items
close to some actual item.  We show empirically that this
approach achieves comparable performance to
state-of-the-art discrete methods, and also offers several key advantages: (a)
it leverages highly optimized ML software and hardware for ease of
implementation, performance and  parallelization; (b) it generalizes to a
variety of query types, including those involving items not present in the
dataset, or partially specified items; (c) it accommodates continuous item
attributes; and (d) it can use arbitrary differentiable metrics and non-linear
utilities, allowing end-to-end optimization using gradient ascent.

% \begin{enumerate}
%     \item Leveraging highly optimized ML software and hardware for ease of implementation, high performance and easy parallelization.
%     \item Generalizes to a wider variety of query types, including queries about infeasible or partially specified items (see below).
%     \item Trivially generalizes to continuous or infinite item spaces.
%     \item Can be combined with arbitrary differentiable value metrics (not just the expected utility of the top item) and non-linear utility models, allowing end-to-end optimization using gradient descent.
% \end{enumerate}

Our second contribution leverages this flexibility---we propose a novel elicitation
strategy based on \emph{partial comparison queries}. In multi-attribute domains where
items have large numbers of attributes,
asking a user to compare complete items can be impractical (e.g., with voice
interfaces) or cognitively burdensome. Instead we ask the user to compare
partially instantiated items, with only a small number of attributes specified
as is common in decision analysis \cite{keeney-raiffa}. Finding EVOI-optimal 
partial queries is a difficult combinatorial problem, but we develop a simple, efficient 
continuous optimization method using the ideas above that performs well in practice.

% CEB: can be deleted if space is needed
The remainder of the paper is organized as follows. We outline related work 
in Sec.~\ref{sec:related}. In Sec.~\ref{sec:background} we lay out the framework
and preliminary concepts upon which our contributions are based, including:
Bayesian recommendation, recommendation slates, EVOI, preference elicitation, and 
several existing computational approximations used for Bayesian elicitation, including 
the \emph{query iteration} algorithm \cite{viappiani:nips2010}. We introduce our basic
continuous relaxation for Bayesian recommendation and EVOI computation using
choice or comparison queries involving fully specified items in
Sec.~\ref{sec:continuous}, and several algorithmic variants based on it. In
Sec.~\ref{sec:partial} we apply the same principles to elicitation using queries
that use only subsets of attributes. We provide detailed empirical evaluation of
our methods in Sec.~\ref{sec:experiments}, analyzing both recommendation quality as
a function of number of queries, and the computational effectiveness of our methods.
We conclude with brief remarks on future research directions in Sec.~\ref{sec:conclude}.

\section{Related Work}
\label{sec:related}

Preference elicitation has long been used to assess decision-maker preferences
in decision analysis
\cite{keeney-raiffa,hamalainen:smc01,white:ejor03}, marketing science \cite{toubia:jmr2004}
and AI \cite{urszula:aaai00,preference:aaai02}, often exploiting the multi-attribute
nature of actions/decisions. In RSs, elicitation has primarily been studied
in domains where items have explicit attributes 
and the recommendable space has a relatively small numbers of items
(e.g., content-based as opposed to collaborative filtering (CF) RSs),
which makes such settings qualitatively similar to other multi-attribute
settings \cite{pu:AIM2008}.

Most work on elicitation assumes some uncertainty representation over user
preferences or utility functions, some criterion for making
recommendations under such uncertainty, and uses one or more types of query to
reduce that uncertainty. A variety of work uses \emph{strict uncertainty}
in which (the parameters of) a user's utility function are assumed to
lie in some region (e.g., polytope) and queries are used to refine
this region. Various decision criteria can be used to make recommendations
in such models (e.g., using a centroid \cite{toubia:jmr2004}), with \emph{minimax regret}
being one popular criterion in the AI literature \cite{boutilier-minimax:AIJ06}.
% \cite{boutilier:regretSurvey2013}. 
Queries can be
selected based on their ability to reduce the volume or surface of a polytope,
to reduce minimax regret, or other heuristic means. 
%TL: Craig TODO
%In this

A variety of query types can be used to elicit user preferences.
% Ivan: removing next three lines since they are redundant with end of paragraph.
%for example, the pairwise comparison (asking a user which of two items
%she prefers) or the generalization to choice queries (asking which item is
%most preferred from a slate of $k$ options).
It is common in multi-attribute utility models to exploit utility function
structure to ask users about their preferences for small subsets of attributes
\cite{keeney-raiffa,fishburn:ier67,braziunas-gaiElic:2005}. 
Techniques based on multi-attribute utility theory---the focus of our work---have been
applied to RSs (see Chen \& Pu \shortcite{Chen04elicitation} for an overview). 
For example, critiquing methods \cite{burke-critiquing,viappiani:06} present candidate items to
a user, who critiques particular attributes (e.g., ``I'd prefer a smaller item'') to drive
the next recommendations toward a preferred point, while unconstrained natural language conversations
offer further flexibility \cite{radlinski:sdd2019}.
We focus on elicitation methods that use \emph{set-wise/slate choice queries}: a user is presented with a slate of $k$ (often multi-attribute) items and asked to state
which is their most preferred. If $k=2$, this is a classic
\emph{pairwise comparison}. Such queries are common in decision support, conjoint analysis
% , interactive configuration
and related areas. 
% While distribution-free criteria
%  (e.g., minimax regret \cite{viappiani:recsys09}, volumetric heuristics \cite{toubia:jmr2004}) are often used, here we focus on Bayesian techniques.

% For
% instance, critiquing-based RSs allow users to make changes to individual
% item attributes to improve an item's utility \cite{burke-critiquing}.

In this work, we focus on Bayesian elicitation, in which a distribution
over user utility functions (or parameters) is maintained and updated
given user responses to queries.
A fully Bayesian approach requires one to make recommendations using expected
utility w.r.t.\ this distribution \cite{urszula:aaai00,preference:aaai02,viappiani:nips2010},
but other criteria can be used for reasons of computational efficiency,
e.g., 
% optimizing w.r.t.\ a maximum-likelihood estimate of user utility, or 
minimizing maximum expected regret (loss) of a recommendation w.r.t.\
the utility distribution \cite{bourdacheEtAL:ijcai19}.

\emph{Expected value of information} (EVOI) \cite{howard:readings} provides
a principled technique for elicitation in Bayesian settings. EVOI requires
that one ask queries such that posterior decision (or recommendation)
quality (i.e., user utility) is maximized in expectation (w.r.t.\ to possible user responses). Such queries are, by construction, directly
relevant to the decision task at hand.
% is often a more useful Bayesian criterion
% where queries are selected to maximize the expected increase in user utility from the information gathered by asking the query, hence focusing on queries that provide decision-relevant information. 
In RSs, ``value'' is usually defined as the utility of the top recommended item. Guo and Sanner \shortcite{sanner:aistats10} select high EVOI queries assuming a diagonal Gaussian distribution over user utilities, but their algorithms evaluate all $O(n^2)$ pairwise queries (or $O(n)$ in their ``greedy'' algorithm), neither of which are practical for real-time elicitation over millions of items. The most direct predecessor of our work is that of Viappiani and Boutilier \shortcite{viappiani:nips2010}, who propose an approximate iterative algorithm for EVOI optimization that only considers a small number of queries. We provide details of this approach in Sec.~\ref{sec:queryiteration}.
% Ivan: commenting next lines out because I don't think we've shown it.
%However, this approach
%is best-suited to content-based RSs with relatively small numbers of items.

Other approaches to elicitation using distributions over user preferences
include various forms of \emph{maximum information gain}.
These ask the query whose expected
response offers maximal information according to some measure. 
Canal et al.~\shortcite{canal_active:2018} select pairwise comparisons by maximizing mutual information between the user response and the user preference point in some high-dimensional space. Rokach et al.~\shortcite{Rokach:2012} select the pairwise query that minimizes ``weighted generalized variance,'' a measure of posterior spread once the query is answered, while Zhao et al.~\shortcite{zhao2018cost}
ask the ``least certain'' pairwise comparison constructed from the top $k$ 
items (or between the top $k$ and the rest). While often more tractable
than EVOI, information gain 
criteria suffer (compared to EVOI) from
the fact that not all information has the same influence on recommendation quality---often only a small amount of information is decision-relevant.
% (e.g., learning the user's relative preference between her 100th and 101st favorite items is irrelevant in most cases).
Other non-Bayesian query selection criteria are possible; e.g.,
Bourdache et al.~\shortcite{bourdacheEtAL:ijcai19} use a variant
of the current solution strategy \cite{boutilier-minimax:AIJ06}
for generating pairwise comparison queries.

The continuous optimization method we propose in
this work can handle arbitrary combinatorial items with both discrete and continuous attributes. Such a setting is studied in \cite{dragone2018}, 
% who take the unusual approach of not maintaining a probability distribution over user utilities, and instead maintain 
where point estimates of utility are used, with a structured perceptron-like update after each query.
% Instead of maximizing EVOI, heuristic query selection is tailored to take advantage of this update rule.
Dragone et al.~\shortcite{dragone2018decomposition} use partially specified items for elicitation, but consider a critiquing model of interaction whereas we focus on set-wise comparisons.
% since they require less effort and domain knowledge on the part of the user.

While content-based RSs model user preferences w.r.t.\ item
attributes, in large, commercial RSs
user preferences are usually estimated using
user interaction data collected passively. Preferences are modeled
using CF \cite{grouplens:cacm97,SMBPMF},
neural CF \cite{yiEtAl:recsys19}) or related models, which
often construct their own representations of items, e.g., by embedding
items in some latent space. 
%This makes elicitation challenging
%conceptually and computationally.  However, elicitation has been 
%investigated in such setting.
Among the earliest approaches to elicitation with latent item embeddings are active CF techniques 
\cite{grouplens:active2002,activecf:uai03} that explicitly query
users for item ratings; more recently Canal et al.~\shortcite{canal_active:2018} similarly elicit over item embeddings. In this work, we also generate comparison
queries using learned item embeddings, though our methods apply equally
to observable item attributes.

% One intuitive criterion for \emph{minimax regret}, as used in \cite{viappiani:2013}; queries are selected so as to maximize worst case utility. While useful in some high-stakes settings, minimax-regret and other non-probabilistic criteria don't allow us to take advantage of previous user interactions to learn a prior over preferences, and regret-based methods are generally not robust to noisy and inconsistent user feedback.

% The above list of criteria for elicitation is far from exhaustive; there are a number of other approaches in the literature, including \cite{Christakopoulou:2016} who focus on candidate/critique interactions but also consider pairwise comparisons which they select using a greedy algorithm inspired by dueling bandits, and  \cite{Sepliarskaia:2018}, who select queries to minimize binary classification error. 

Finally, we note that elicitation has strong connections to work on Bayesian
(or blackbox) optimization \cite{bayesian_optimization_survey:ieee15}, where the aim is to find the optimum of some 
function---analogous to finding a good recommendation for a user---while
minimizing the number of (expensive) function evaluations---analogous to minimizing the number/cost of user queries. In contrast with most work in
preference elicitation, Bayesian optimization methods typically focus
on direct function evaluation, equivalent to asking a user for their
utility for an item, as opposed to asking for a qualitative
comparisons, though recent work considers dueling bandit models using
pairwise comparisons in Bayesian optimization \cite{gonzalezEtAl:icml17,suiEtAl:ijcai18}. Such techniques are generally derivative-free, though recent
work considers augmenting the search for an optimum with (say, sampled) gradient
information (e.g., \cite{wuEtAl:nips17}). We use gradient information when
computing EVOI, directly exploiting the linear nature of user
utility in embedding space.

\section{Background}
\label{sec:background}

We begin by outlining the basic framework, notation, and prior results on which our methods are based.

\subsection{Bayesian Recommendation}
\label{sec:bayesrec}

We adopt a Bayesian approach to recommendation and elicitation in which the RS
maintains a \emph{probabilistic belief} about a user's utility function over recommendable
items. It uses this to generate both recommendations and elicitation queries.
A \emph{recommendation problem} has six main components: an \emph{item set}, a user 
\emph{utility function} drawn from a utility space, a \emph{prior belief} over utility space,
a \emph{query space}, a \emph{response space}, and a \emph{response model}.
We outline each in turn.

We assume a set of $N$ 
recommendable \emph{items} $\calX \subseteq \R^d$, each an instantiation of $d$ (for ease of
exposition, real-valued) attributes (categorical attributes can be converted in
standard fashion). These  attributes may be dimensions in some latent space, say, as generated by some neural CF model (see Sec.~\ref{sec:experiments}.) A \emph{user}, for whom a recommendation is
to be made, has a \emph{linear utility function} $u$ over items, parameterized as a vector $\bfu \in \calU$, where $\calU \subseteq \R^d$; i.e., $u(\bfx ; \bfu) = \bfx^T \bfu$ for any $\bfx \in \calX$.\footnote{Linearity of utility may be restrictive if the
attributes are observable or ``catalog'' properties of items as opposed to reflecting
some learned embedding. Our methods can be extended to other utility representations in
such a case, for example, UCP- \cite{ucpnets:uai01} or GAI-networks \cite{gonzales:kr04,braziunas-gaiElic:2005}, which offer
a linear parameterization of utility functions without imposing linearity w.r.t.\  the
attributes themselves.} A user's most
preferred item is that with greatest utility: 
% We focus single recommendations, where the user is interested in consuming the item with highest utility:
$
\bfx^*_{\bfu} = \argmax_{\bfx} u(\bfx ; \bfu).
$
The \emph{regret} of any recommendation $\bfx\in\calX$ is
$
    \Regret(\bfx ;\bfu) = u(\bfx ; \bfu) - u(\bfx^*_\bfu ; \bfu), 
$
and is a natural measure of recommendation quality. Finally, we assume the RS has some
prior belief $P_0$ over $\calU$. This reflects its uncertainty over the true utility function $u$ of the user.

The prior $P_0$ is typically derived from past interactions with other users, and reflects the heterogeneity
of user preferences in the domain. While we explicate our techniques in the \emph{cold-start}
regime where the RS has no information about the user in question, $P_0$ may also incorporate past interactions
or other information about that user---this has no impact on our methodology. The prior will be updated as the
RS interacts with the user, and we use $P$ generically to denote the RS's current belief.

Given a belief $P$, the \emph{expected utility} of an item (or recommendation) $\bfx$ is:
\begin{equation}
    \eu(\bfx;\! P) \! =\! \Exp_{P} 
      \left[ \bfx^T \!\bfu \right] 
      \! =\! \bfx^T\!\! \left[ \Exp_{P} \bfu \right]
      \! =\! \bfx^T\!\! \left[\int_{\calU} \bfu P(\bfu) d\bfu \right]\! .
\end{equation}
% 
% \begin{equation}
%     \eu(\bfx ; \bfr) = \Exp_{P(\bfu | \bfr ;\bfq)} 
%       \left[ \bfx^T \bfu \right] 
%       = \bfx^T \left[ \Exp_{P(\bfu | \bfr ;\bfq)} \bfu \right] \; .
% \end{equation}
The optimal recommendation given $P$ is
% during user interaction by selecting an
the item with maximum expected utility:
\begin{equation}
    \bfx^*_{P} = \argmax_{\bfx\in\calX} \eu(\bfx ; P), \;\;\; \eus(P) = \eu(\bfx^*_{P} ; P).
\end{equation}
% \begin{equation}
%     \bfx^*_{\bfr} = \argmax_{\bfx\in\calX} \eu(\bfx ; \bfr), \quad \eus(\bfr) = \eu(\bfx^*_{\bfr} ; \bfr) \; .
% \end{equation}

% We take a Bayesian approach towards recommendations and
% preference elicitation.

% We assume that a \emph{cold-start} user is drawn from a 
% \emph{populational prior},
% $P(\bfu)$ over $\calU$, which the recommender system has access to.
% This could have been estimated from previous user choice datasets.
% Such a prior can reflect the heterogeneity
% of user preference types.
%
The RS can refine its belief and improve recommendation quality by asking
the user questions about her preferences.
% engages with the user in a sequence of queries and responses.
Let $\qs$ be the \emph{query space}. For any query $q\in \qs$,
the \emph{response space} $\rs_q$ reflects possible user
responses to $q$.
% CEB: if room:
For example, a pairwise comparison query (e.g., ``do you prefer $\bfx_1$ to $\bfx_2$?'')
has a binary response space (yes, no).
For any sequence of queries $\bfq = (q_1, \ldots, q_n)$, $n\geq 0$,
to simplify notation
we assume that
any corresponding sequence of user responses
$\bfr = (r_1, \ldots, r_n)$, where $r_i \in \rs_{q_i}$, uniquely determines $\bfq$
(e.g., through suitable relabeling so that a ``yes'' response to some $q$ is
encoded differently than a ``yes'' to a different $q'$).
% the \emph{history} of user preferences. If $n=0$, we have an empty history, $\bfr = \emptyset$.
We also assume the RS has a \emph{response model} that specifies the probability
$R(r_i | q; \bfu)$ of a user with utility function $\bfu$ offering response $r\in\rs_q$ when asked query $q$.
% We discuss specific query/response spaces and response models below.

We focus on \emph{slate (comparison) queries}, $q = (\bfx_1,\ldots, \bfx_k)$,  $k\ge 2$, 
in which a slate of $k$ items is
presented to the user, who is asked which is most preferred.\footnote{The response set $\rs_q = \{\bfx_1, \ldots, \bfx_k\}$ can be augmented
with a ``null'' item to account for, say, a ``none of the above'' response.}
%When $k=2$, this is a classic \emph{pairwise comparison}.
% Partial comparisons are discussed in Sec.~\ref{sec:partial}.
% 
%We make the assumption that user responses
%are independent, i.e.,
%$P(\bfr ~|~ \bfu ; \bfq) = \prod_{i=1}^n P(r_i ~|~ \bfu ; q_i)$. It %is indeed a simplifying assumption but one
%that is common in the elicitation literature 
%(TODO: give some refs?). 
We consider two response models for slate queries. The \emph{noiseless response model} assumes a user
responds with the utility maximizing item: $R(\bfx_i | q; \bfu) = \1[i = \argmax_{j=1}^k u(\bfx_j ; \bfu)]$.
The \emph{logistic response model}
% CEB: add back if space allows
% (a.k.a.\ Luce-Sheppard or Bradley-Terry in the pairwise case) 
assumes 
$R(\bfx_i | q; \bfu) = e^{u(\bfx_i ; \bfu)/\tau} / \sum_{j=1}^k e^{u(\bfx_j ; \bfu)/\tau}$,
where  temperature $\tau > 0$ controls the degree of stochasticity/noise in the user's choice.
Other choice models could be adopted as well \cite{louviere-et-al:statedchoice2000}.

We assume the response to any
query is conditionally independent of any other given $\bfu$.
 Let $R(r | q; P) = \Exp_{\bfu\sim P} R(r | q; \bfu)$ be the expected probability of
 response $r$ given belief $P$.
Given any response sequence $\bfr$ (which determines the generating query sequence $\bfq$)
% from which is was induced)
and current belief $P$, the \emph{posterior belief} of the RS is given by Bayes rule:
$$
   P_\bfr(\bfu) = P(\bfu | \bfr) \propto R(\bfr | \bfq; P) = R(\bfr | \bfq; \bfu) P(\bfu).
$$

\subsection{Expected Value of Information}
\label{sec:evoi}

While computing optimal query strategies can be cast as a sequential decision
problem or POMDP \cite{preference:aaai02,white:ejor03}, we adopt a simpler,
commonly used approach, namely, \emph{myopic} selection of queries using the well-founded
\emph{expected value of information} (EVOI) criterion \cite{howard:readings,urszula:aaai00}.
Given belief $P$, we first define
the expected utility (w.r.t.\ possible responses) of the best recommendation
after the user answers query $q$:
\begin{definition}
The \emph{posterior expected utility} (PEU) of $q$ is:
% given a history $(\bfq, \bfr)$ of queries and user responses is:
% \begin{equation}
%     \peu(q) = \sum_{r\in \rs_q} P_r(r ; q) \eus((\bfr, r)) \; .
% \end{equation}
\begin{equation}
    \peu(q; P) = \sum_{r\in \rs_q} R(r | q; P) \eus(P_r).
\end{equation}
The \emph{(myopic) expected value of information} is 
$\evoi(q; P) = \peu(q; P) - \eus(P)$.
%, a constant offset of PEU.
\end{definition}
A query with maximum PEU maximizes the
expected utility of the best recommendation conditioned on the user's
response. EVOI, which is maximized by the same query, measures the \emph{improvement} in expected utility offered
by the query relative to the prior belief $P$ and can serve as a useful metric for terminating elicitation.

\subsection{EVOI Optimization with Particle Filtering}
\label{sec:particles}

We use a Monte Carlo approach to optimize EVOI
as in \cite{viappiani:nips2010}. Given belief $P$,
we sample $m$ points $\bfU = (\bfu_1, \ldots, \bfu_m)$ from
$P$ and maximize the sample average to
approximate the integral within EVOI:
% Without loss of generality, assume we have a 
% posterior $P$ conditioned on previous history $\bfq$ and $\bfr$.
% We have,
\begin{align*}
\peu(q; P) &= \sum_{r\in \rs_q} R(r|q;P) \max_{\bfy_r \in \calX}
  \left[ 
\int_{\bfu} u(\bfy_r ; \bfu) P_r(\bfu) d\bfu \right] \\
%% &= \sum_{r\in \rs_q} R(r|q;P) \max_{\bfy_r \in \calX} \left[ \int \bfy_r^T 
%% \bfu \frac{R(r | q; \bfu) P(\bfu)}{R(r|q;P)} d\bfu \right] \\
&\approx \sum_{r\in \rs_q} \max_{\bfy_r \in \calX} \left[ \frac{1}{m}
  \sum_{j=1}^m \bfy_r^T \bfu_j R(r|q;\bfu_j) \right].
\end{align*}
% Note that any query which maximizes EVOI also maximizes PEU. 
For slate queries under logistic response, an optimal
query $q^\ast = (\bfx_1, \ldots, \bfx_k)$ w.r.t.\ EVOI satisfies:
\begin{multline}
    % \argmax_q \evoi(q) = \argmax_q \peu(q) \\
    % = \argmax_{\bfx_1, \ldots, \bfx_k} \sum_{i=1}^k 
    \max_{\bfx_1, \ldots, \bfx_k} \sum_{i=1}^k 
      \max_{\bfy_i \in \calX} \bfy_i^T \!\! \left[\! \frac{1}{m}
      \sum_{j=1}^m \bfu_j \frac{e^{\bfx_i^T \bfu_j/\tau }}{
      \sum_{\ell=1}^k e^{\bfx_\ell^T \bfu_j / \tau}} \!\right].
      \label{eq:max_evoi}
\end{multline}
Computing EVOI for a single query using this approach
% takes time 
requires $O(N mk^2)$ time.
% CEB: took out for space, but could go back in
% ---an $O(N)$ scan for the best $\bfy_i$ and
% $O(mk)$ for computing the empirical average utility parameters (inner
% square brackets), repeated $k$ times.
%
Consequently, 
% exhaustive
search over all $\binom{N}{k}$ queries to maximize EVOI is prohibitively expensive, even in the pairwise case ($k=2$), when $N$
is moderately large.

\subsection{Query Iteration}
\label{sec:queryiteration}

While finding optimal EVOI queries is intractable, effective heuristics are available.
Notice that computing EVOI for a query $q = (\bfx_1 \ldots \bfx_k)$ requires identifying the items with greatest
\emph{posterior}
expected utility conditioned on each potential user response $\bfx_i$, i.e., the maximizing items $\bfy^*_i$ in Eq.~\ref{eq:max_evoi}. We refer to this operation as a \emph{deep retrieval} for query $q$, and write $(\bfy^*_1\ldots\bfy^*_k) = \DR(\bfx_1 \ldots \bfx_k)$. We
sometimes refer to $(\bfx_1\ldots\bfx_k)$ as the \emph{query slate} and $(\bfy^*_1 \ldots \bfy^*_k)$
as the \emph{(posterior) recommendation slate}.

% Considering the $\bfy_i$'s in Eq. \ref{eq:max_evoi} above.

% One of the practical bottlenecks
% when $\calX$ is large is finding the maximizing $\bfy_i$.
% %
% \begin{definition}
% Given $\bfv \in \R^d$, the \emph{deep retrieval problem} is that of finding
% the maximizer $\DR(\bfv) := \argmax_{\bfy \in \calX} \bfy^T \bfv$. In particular, the
% \emph{deep retrieval problem given a user response} of selecting item
% $\bfx_i$ to a query $q=(\bfx_1,\ldots\bfx_k)$ is 
% $\DR(r=\bfx_i, q) := \argmax_{\bfy \in \calX} \eu(\bfy ; r=\bfx_i) = \argmax_{\bfy\in\calX} \bfy^T \sum_{j=1}^m \bfu_j 
% \left[\nicefrac{e^{\tau \bfx_i^T \bfu_j}}{\sum_{\ell=1}^k e^{\tau \bfx_\ell^T \bfu_j}}\right]$.
% \end{definition}

The following result tells us that replacing the query slate with the induced recommendation slate increases EVOI:
\begin{thm}[\citeauthor{viappiani:nips2010} \citeyear{viappiani:nips2010}]
\label{thm:viappiani}
Let $q=(\bfx_1, \ldots, \bfx_k)$ be a slate query
%(maximizes Eq.~\ref{eq:max_evoi})
with $q' = (\bfy^*_1\ldots\bf^*y_k) = \DR(q)$.
% $\bfy^*_i = \DR(r=\bfx_i, q)$ for  $i\le k$.
% $\bfx_i$ by $\bfy^*_i$
Under noiseless response, $\evoi(q';P)\geq \evoi(q;P)$; while under
logistic response, $\evoi(q';P)\geq \evoi(q;P) - \Delta$, where 
$$\Delta\!=\! \sum_{i=1}^k \left[\!\int\! \left(
             1 - 
              % \nicefrac{e^{\bfx^T_i \bfu/\tau }}              
              \frac{e^{\bfx^T_i \bfu/\tau }}
             {\sum_{\ell=1}^k e^{\bfx_\ell^T \bfu/ \tau }}\right) P_{{\bfx_i}}\! (\bfu)d\bfu \right]
             \eu\! (\bfy^*_i; P_{{\bfx_i}}\! ).
$$
% \begin{enumerate}
%     \item[1)] cannot decrease under hardmax response, and
%     \item[2)] can decrease by at most 
%     $\sum_{i=1}^k [\int(
%              1 - 
%               \nicefrac{e^{\bfx^T_i \bfu/\tau }}
%              {\sum_{\ell=1}^k e^{\bfx_\ell^T \bfu/ \tau }}) P(\bfu | r; q)d\bfu ]
%              \eu(\bfy^*_i ; r=\bfx_i)$ under softmax response.
% \end{enumerate}
\end{thm}
%\begin{corollary}\label{cor:feasibleopt}
%Under hardmax, there always exists a slate %query with feasible items
%from $\calX$ that maximizes EVOI.
%\end{corollary}
% Therefore under hardmax response or softmax with sufficiently large $\tau$,
% an optimal slate query will have the property that after a user
% selects item $\bfx_i$, the best resulting recommendation will
% also be $\bfx_i$ (w.r.t. posterior utility model).
This suggests a natural iterative
algorithm $\qi$: start with a query slate, replace it with its induced recommendation slate, repeat until convergence. Viappiani and Boutilier \shortcite{viappiani:nips2010} find that, in practice, $\qi$ converges quickly and finds high EVOI queries in settings with small item sets.

\section{Continuous EVOI Optimization}
\label{sec:continuous}

While $\qi$ is often an effective heuristic, it cannot scale to
settings with large item spaces: each iteration requires the computation of the
item with maximum PEU for each of the $k$ responses; but computing this maximum 
generally
requires EU computation for each candidate item.

To overcome this, we develop a continuous optimization formulation for EVOI maximization. Intuitively, we ``relax'' the discrete item space $\calX$ to obviate
the need for enumeration, and treat the items in the query and recommendation
slates as continuous vectors in $\R^d$. Once EVOI is optimized in the relaxed problem
using gradient ascent, we project the solution
back into the discrete space $\calX$ to obtain a slate query using only feasible items.
Apart from scalability, this offers several advantages: 
we can exploit highly optimized and parallelizable 
ML frameworks (e.g. TensorFlow) and hardware for
ease of implementation and performance;
variants of our methods described below have common, reusable computational
elements; and the methods are easily adapted to continuous item spaces and novel query types (see  Sec.~\ref{sec:partial}.)
%
% As mentioned earlier, a continuous approach has a number of potential advantages over query iteration and similar algorithms.
% We can piggyback on highly optimized and parallelizable 
% ML frameworks and hardware both for
% for ease of implementation and performance. And as we show, the
% variants of our continuous methods have common re-usable elements, and readily adapt
% to a wider variety of queries (e.g. Sec.~\ref{sec:partial}),
% and we can leverage multiple gradient-based
% optimizers as long as our formulations are compositions of
% differentiable functions.
% COMMENT(Ivan): don't think we should claim this here since we have no experiments showing it...

\subsection{A Reference Algorithm}

We develop a basic continuous formulation by considering the EVOI objective
in Eq.~\ref{eq:max_evoi}---we focus on logistic response for concreteness. In the discrete case, each item $\bfx_i$ (in the query slate) and $\bfy^*_i$ (recommendation slate) are optimized by enumerating the feasible item set $\calX$.
We relax the requirement that items lie in $\calX$ and treat these as continuous variables. Let $\bfX$ (query slate) and $\bfY$ (recommendation slate) be two $d\times k$ matrices whose $i$-th columns represent $\bfx_i$ and $\bfy^*_i$, respectively.
Let $\bfU$ be an $m\times d$ matrix whose rows are the sampled utilities
$\bfu_i$.
We can express the softmax probabilities in the inner
sum of Eq.~\ref{eq:max_evoi} as 
$\mathrm{softmax} (\bfU \bfX)$, constructing a row vector of probabilities
(each element is a logit).

Similarly, the dot products ${\bfy^*_i}^T \bfu_j$ in the outer sum can be 
expressed as $\bfU \bfY$. The Hadamard (element-wise) product gives:
$$
[\bfU\bfY\circ \mathrm{softmax} (\bfU \bfX)]_{ij} = 
{\bfy^*_j}^T \bfu_i  \frac{e^{\bfx_j^T \bfu_i/\tau }}{\sum_{\ell=1}^k e^{\bfx_\ell^T \bfu_i / \tau}}.
$$
Summing over $j$ and averaging over $i$, we obtain:
\begin{multline}
\frac{1}{m}\1_m^T \bfU\bfY\circ \mathrm{softmax} (\bfU \bfX) \1_k = \\ \frac{1}{m}\sum_{i=1}^m\sum_{j=1}^k {\bfy^*_i}^T \bfu_j \frac{e^{\bfx_j^T \bfu_i/\tau }}{\sum_{\ell=1}^k e^{\bfx_\ell^T \bfu_i / \tau}} , \label{eq:matrix_max_evoi}
\end{multline}
where $\1_d$ is a $d$-vector of 1s.
If we maximize $\bfY$ for any given $\bfX$, this is exactly the PEU of query
slate $\bfX$.
% As we discuss below this decomposition into query and rec. slate will play an important role in our algorithms.
We can then apply gradient-based methods to optimize $\bfY$ (to compute PEU)
and $\bfX$ (to find the best query) as we detail below.
% CEB: not needed here
% (perhaps subject to some norm bounds).\footnote{We can 
% normalize items to zero mean and identity covariance.}

A solution $\bfX$ may contain infeasible items depending on the optimization
method used, in which case we must project the
query slate $\bfX$ into item space $\calX$---we discuss mechanisms for this below.
One approach is to leverage Thm.~\ref{thm:viappiani}, and perform one deep retrieval to select feasible (optimal) recommendation items $\bfy^*_i$ given $\bfX$. Thm.~\ref{thm:viappiani} ensures that $\{\bfy^*_i\}$, interpreted as a query, is 
a better feasible slate (or at least not much worse) than $\bfX$. To avoid duplicate $\bfy^*_i$'s, we deep retrieve the top-$k$ items for each possible user response and only add distinct items to the query slate (see Alg.~\ref{alg:drtopk}). Alg.~\ref{alg:contbase} details this basic
``reference'' algorithm. We now discuss variations of the reference algorithm.
%
% DEEP RETRIEVE TOP-K
%
\begin{algorithm}[t]
 \caption{$\druniq$. Inputs: optimized $\bfX^*$ and $\bfU$} 
 \label{alg:drtopk}
 \begin{algorithmic}[1]
   \STATE Let $\calX$ be an $N\times d$ matrix of all feasible items.
   \STATE $S \leftarrow \emptyset$
   \STATE $\bfp = \mathrm{softmax}(\bfU\bfX^*)$
   \STATE Let $\bfT$ be a $k\times k$ matrix where the $ij$-th entry is the index of the $j$-th highest element in the $i$-th row of $\bfp^T \bfU \calX^T$
   \FOR{$i=1..k$}
     \FOR{$j=1..k$}
        \IF{$T_{ij} \not\in S$}
            \STATE $S \leftarrow S \cup \{T_{ij}\}$
            \STATE break
        \ENDIF
    \ENDFOR
            
%and break if $T_{ij} \not\in S$.
   \ENDFOR
   \RETURN slate of items indexed by $S$
 \end{algorithmic}
\end{algorithm}
%
% BASELINE (REFERENCE) CONTINUOUS EVOI
%
\begin{algorithm}[t]
 \caption{Continuous EVOI (reference algorithm)}
 \label{alg:contbase}
 \begin{algorithmic}[1]
   \STATE $\bfX^*, \bfY^* \leftarrow \max_{\bfX, \bfY} \1^T_m\bfU \bfY\circ \mathrm{softmax}(\bfU\bfX)\1_k$ using gradient methods.
   \RETURN $\druniq(\bfX^*, \bfU)$ as query slate
 \end{algorithmic}
\end{algorithm}
%
% NETWORK ARCHITECTURE
%
\begin{figure}[t]%[ht]
\begin{center}
\centerline{\includegraphics[width=0.9\columnwidth]{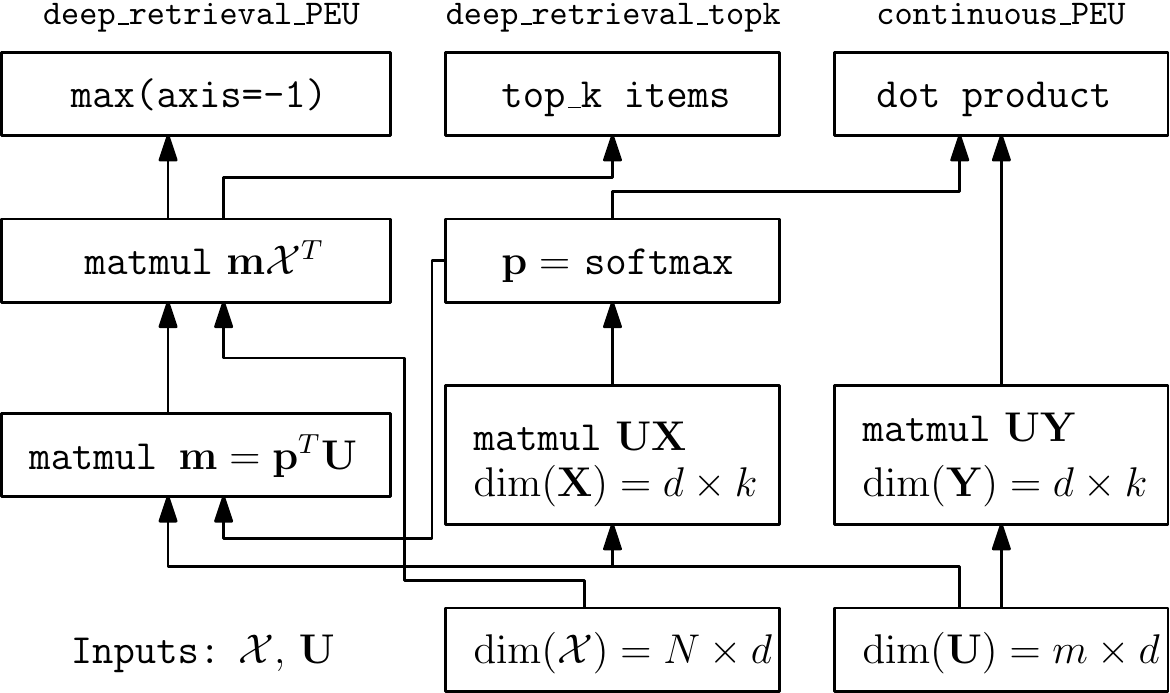}}
    \caption{A common network architecture. % common to our variants of  continuous optimization.
    Input items $\calX$, utility particles 
    $\bfU$ are passed
    through dense layers with weights $\bfX$ (columns are query items) and $\bfY$ (columns are rec.\ items
    \emph{post}-user response). Output {\scriptsize {\tt continuous\_PEU}} is
    PEU with rec.\ slate $\bfY$. If
    the slate is constrained to be feasible,
    {\scriptsize{\tt deep\_retrieval\_PEU}} gives feasible PEU.
    % (though slate $\bfX$ may be infeasible).
    {\scriptsize{\tt deep\_retrieval\_topk}} outputs
     top-$k$ EU items for each response 
    to generate a query slate in $\druniq$.}
\label{fig:slatenet}
\end{center}
\end{figure}

\subsection{Taxonomy of Continuous EVOI Algorithms}

The reference algorithm can be instantiated in different ways,
depending on how we represent the (continuous) query and recommendation slates $\bfX$ and $\bfY$, and how we project these into feasible item space. These three 
choices form a taxonomy of continuous EVOI algorithms:
%
% COMMENT(Ivan): can we remove the figure now that we've explained recommendation slates better above?
% \begin{figure}[t]
% \centerline{\includegraphics[width=0.99\columnwidth]{evoi_decom.pdf}}
% \caption{EVOI decomposition consists of a \emph{query 
% slate} $\{\bfx_i \}_{i=1}^k$ and a \emph{recommendation (rec.)
% slate} $\{\bfy_i \}_{i=1}^k$. Each $\bfy_i$ corresponds to an
% item rec. given the user selects $\bfx_i$ from the query,
% we seek to optimize the sum over the EU of $\bfy_i$ 
% conditioned on the user's choice. We can represent the
% query or rec. items as continuous variables, perhaps
% with feasibility constraints as soft penalties, 
% etc. After optimizing, we must project variables into
% feasible item space.\label{fig:evoi_decomp}}
% \end{figure}

{\flushleft \textbf{Query Slate Representation}}.
The simplest way to represent queries are as ``unconstrained'' 
variables $\bfx_i \in \R^d$ uninfluenced by the feasible set $\calX$ (though
possibly with some norm constraint).
However, in most applications feasible items will be sparsely
distributed in $\R^d$ and such unconstrained optimization may yield queries far from any feasible item. 
To address this, we can incorporate ``soft'' feasibility constraints as 
regularization terms in the optimization, encouraging each $\bfx_i$ 
to be near some feasible item. We note that this restriction can magnify local optimality issues.

{\flushleft \textbf{Recommendation Slate Representation}}.
As above, we can also treat the recommendation slate items $\bfy^*_i$ as unconstrained variables or regularize them to be near feasible items. We consider two
other approaches, allowing a tradeoff between computational ease and fidelity of EVOI computation. The first \emph{equates}
the query and recommendation slates.
% 
% {\it Equate Query and Recommndation Slates}. 
Thm.~\ref{thm:viappiani} 
implies that there exist optimal EVOI queries where 
% the query slate and the recommendation slate are one and the same, 
the two coincide, i.e. $\bfy^*_i = \bfx_i$, under noiseless response, and do so approximately
under logistic response.
% While this result only holds in softmax response models up to some error
% bounds, we can use it as a heuristic to simplify our 
Using a single set of slate variables in Eq.~\ref{eq:matrix_max_evoi} for both gives the following optimization:
% by equating the query slate and and recommendation slate variables.
% Eq.~\ref{eq:matrix_max_evoi} then simplifies to 
\begin{small}
\begin{equation}
    \argmax_{q}\! \evoi(q)\! \approx\! \argmax_{\bfX}\! \frac{1}{m}\1_m^T\! \bfU\bfX\circ \mathrm{softmax} (\! \bfU \bfX\! ) \1_k .
             \label{eq:freeevoi}
\end{equation}
\end{small}
% {\it Deep Retrieval}. 
A more computationally expensive, but more
accurate, approach for EVOI avoids relaxing the query slate, instead
% representing the items entirely by 
computing exact EVOI for $\bfX$. This requires $k$ deep retrieval operations
(one per item) over $\calX$
% ($k$ max ops over feasible items)
at each optimization step. In practice, we can apply deep 
retrieval only periodically, giving an alternating optimization procedure that 
% EM-like optimization, 
optimizes query variables given the recommendation slate at one stage and
deep retrieving the recommendation items given the query variables the next.
One could also compute the max over a small subset of prototype items.

{\flushleft \textbf{Projecting to Feasible Items}}.
The $\bfx_i$ (query) values after optimization will not generally
represent feasible items. We
could recover a feasible 
query by projecting each $\bfx_i$ to its nearest neighbour in $\calX$, 
but this might give poor queries, e.g.,
% CEB: This is a very awkward sentence to end on. Then what?
by projecting an informative infeasible pairwise comparison 
to two feasible items that differ in only a dimension 
that offers no useful information.
% (e.g., a user will always prefer cheaper item when all other attributes are equal).
Instead, we use a single deep retrieval of $\bfX$ to
obtain a set of $\bfy^*_i$ to serve as the query.

\subsection{Variants of Continuous EVOI Algorithms}

We identify four natural algorithms motivated by the
taxonomy above. The modularity of our approach allows the design of a single network architecture, shown in Fig.~\ref{fig:slatenet},
that supports all of these algorithms due to its reusable elements. See Appendix~\ref{appendix:algos} for additional details.
\begin{itemize}
\item $\freeevoi$. Equate query and recommendation slates; unconstrained except
for an $\ell_2$ norm bound on variables.
\item $\regevoi$. Like $\freeevoi$ but with an added
$\ell_2$ regularization
$\lambda \sum_{i=1}^k \min_{\bfz \in \calX} \| \bfx_i - \bfz \|^2_2$ 
ensuring the variables stay close to feasible items.
\item $\drevoi$. Query variables are unconstrained,  deep retrieval occurs at every optimization step.
\item $\alterevoi$. Query variables are unconstrained, and we use the alternating optimization described above.
\end{itemize}
All four algorithms use one deep retrieval at completion
% at the end of the optimization
to ensure the resulting slate query uses feasible items in $\calX$.
% represent feasible queries; but in Sec. 
% \ref{sec:partial} below we describe the partial comparison setting 
% where this is not possible, and our proposed algorithm projects to the 
% nearest feasible queries instead.
%

\subsection{Initialization Strategies}

Since the optimization objective is non-convex,
we can (and do, in practice) encounter multiple local optima in each algorithm
variant. Hence, the initialization
strategy can impact results, so we consider several possibilities.
The first strategy, $\random$, initializes the optimization with random vectors.
The second follows \cite{viappiani:nips2010}, initializing the slate query with the highest-utility items for $k$ randomly selected utility particles
(which they find performs
well in conjunction with $\qi$). We call this strategy
$\paolo$.
A third strategy, $\balanced$, initializes using a query slate
that splits the sampled utility particles into $k$ evenly
weighted groups in which each of the $k$ responses
results in roughly $m/k$ utility particles that are 
(mostly) consistent with the response. % (via soft- or hardmax).
Such balanced queries may
use infeasible items.
% ($\qi$ does not use this strategy).
In practice, we use multiple restarts to mitigate the effects of local maxima.

\section{Partial Comparison Queries}
\label{sec:partial}

In many domains, slate queries with fully specified items
are impractical: they can impose a significant cognitive
burden on the user (e.g., comparing two car configurations with dozens of attributes);
and often a user lacks the information to determine a preference for
specific items (e.g. comparing two movies the user has not seen).
It is often more natural to use \emph{partial comparison queries}
with items partially specified using a small subset of attributes (e.g.,
truck vs.\ coupe, comedy vs.\ drama). 
Finding partial queries with good EVOI requires searching 
through the combinatorial space of queries (attribute subsets).
% and no existing methods directly address this case.
Unlike full queries, with partial queries
the query and item spaces are
distinct. However, we can readily adapt our continuous EVOI
methods.
% to this setting.

\subsection{Semantics of Partial Comparisons}

%What does it mean, exactly, that the user prefers comedies to dramas? Do they prefer every single comedy over every single drama? The average comedy over the average drama? Comedy over drama all else being equal?

The most compelling semantics for partial comparisons may generally
be domain specific,
% (an empirical question that may be addressed with user studies).
but in this work we adopt the 
simple \emph{ceteris paribus} 
(all else equal) semantics~\cite{fishburn:1977}.
Let $\calX$ be defined over $d$ attributes $A$. Given $S\subseteq A$, a \emph{partial item} $\bfx_S$ is a vector with values defined over attributes in $S$. 
We assume there is some $\bfy \in\calX$ s.t.\  $\bfy_S = \bfx_S$
(i.e., any partial item has a feasible completion).
Given partial query slate $q=( \bfx_{1,S} \ldots \bfx_{k,S} )$,
a \emph{uniform completion} of $q$ is any full query $q' = (\bfx_1\ldots \bfx_k)$ 
such that $\bfx_{1,A\backslash S} = \cdots = \bfx_{k,A\backslash S}$ (i.e., each
item is completed using the same values of attributes $A\backslash S$).
\emph{Ceteris paribus responses} require that, for \emph{any} uniform
completion $q'$ of $q$:
\begin{equation*}
    P_r(\bfx_{i,S} | q; \bfu) = P_r(\bfx_i | \bfu, q') \; .
\end{equation*}
This condition holds under, say, additive independence \cite{fishburn:ier67}
and induces an especially simple response model if we assume utilities are linear.
For instance, with logistic responses, we have:
% Since we restrict to soft- and hardmax response models with linear utilities this probability reduces to a nice form:
\begin{equation*}
P_r(\bfx_{i,S} | q; \bfu) = \mathrm{softmax}(\bfu_S^T \bfx_{1,S}, \ldots, \bfu_S^T \bfx_{k,S}) \; .
\end{equation*}
%Let $Y_1$, $Y_2 \subset A$ be two partially specified items. Then we define the \emph{ceteris paribus} semantics to mean that for all realizations of attribute values $Z \subseteq A\backslash (Y_1 \cup Y_2)$ outside $Y_1$ and $Y_2$, the probability of the user picking $Y_1$ over $Y_2$ is
%\begin{multline}
%P(r = Y_1 | \bfu, q=\{Y_1, Y_2\}) \\
%= P(r=X \cup Y_1 | \bfu, q=\{X \cup Y_1, X %\cup Y_2\} \\
%= \textrm{softmax}(u(X \cup Y_1; \bfu), u(X %\cup Y_2; \bfu)) \\
%= \textrm{softmax}(u(Y_1; \bfu), u(Y_2; %\bfu)) \\
%\end{multline}
%
%by linearity of $u$ and the fact that softmax is invariant to subtracting a constant offset $u(X; \bfu)$, and this generalizes to queries comparing $k > 2$ partial items.
Thus our response model for partial queries is similar to 
those
% the soft/hardmax model we use 
for full queries. The optimal EVOI problem must find both attributes and partial items:
% (EVOI defined as in Eq. \ref{eq:max_evoi})
$$
\argmax_{S, \bfx_{1,S} \ldots \bfx_{k,S}} \evoi(\bfx_{1,S} \ldots \bfx_{k,S}) \; .
$$

%In practice we want to reduce user cognitive load by only asking about a small number of attributes $p$ for each item, so we constrain each $\bfx_{i,S}$ to have at most 
%so each $Y_i$ actually only ranges over $p$-subsets of A.

\subsection{Continuous EVOI for Partial Comparisons}

% As described in the taxonomy of continuous EVOI algorithms, 
As in Sec.~\ref{sec:continuous}, we need a continuous slate representation and 
% a method for projecting back to feasible items post-optimization.
projection method.
For query representation, we relax partial items to fall outside $\calX_{S}$.
% relax discrete attributes to take on continuous values (we can convert 
% categorical attributes to binary). 
W.l.o.g., let attribute values be binary with a relaxation in  $[0, 1]$ and 
% for purposes of optimization, 
a partial item vector be a point in $[0,1]^{d}$. Rather than representing the
recommendation slate explicitly, we compute exact EVOI at each step of optimization (deep retrieval). We project to partial queries 
% with low cognitive load
by only specifying a small number of attributes $p$ for each item: we discretize by setting the $p$ highest attribute values for each item to $1$ and the rest to $0$.
This projection could cause arbitrary loss in EVOI, so we use $\ell_1$-regularization to encourage each $\bfx_i$ to have $p$ values close to $1$ and the rest close to $0$. The optimization objective becomes:
\begin{equation}
\argmax_{\bfx_1 \ldots \bfx_k \in [0,1]^{d}} \evoi(\bfx_1\ldots \bfx_k) - \lambda \sum_i \| \textrm{sort}(\bfx_i) - \mathbf{j} \|_1
\label{eq:partial_cont}
\end{equation}
where $\textrm{sort}(\cdot)$ sorts elements of a vector in ascending order, $\mathbf{j}$ is a $d$-dimensional vector with the last $p$ coordinates set to $1$ and the rest to $0$, and $\lambda$ is the regularization constant. We call this the $\contpartial$ algorithm.

\section{Experiments}
\label{sec:experiments}
%
%CEB: STATE THE PURPOSE OF THE EXPERIMENTS UP FRONT: WHAT WILL YOU COMPARE, WHY? ALSO FOR
%EACH DATA SET SUBSECTION. START WITH A *BRIEF* MENTION OF THE DATA SET, THEN POINT TO APPENDIX (LIKE SEC 6.2 DOES).
%

We assess our continuous EVOI algorithms by comparing their speed and the quality of queries they generate to several baseline methods in two sets of experiments: the first uses full-item comparison queries, the second uses our partial comparison queries.
% we introduced in Sec. \ref{sec:partial}.
We are primarily interested in comparing elicitation performance
using (true) regret over
complete elicitation runs---in a real scenario, one would
generally evaluate the quality of recommendations after
each round of elicitation. Since all methods attempt to maximize EVOI,
we also report EVOI.

In all experiments, we sample
utility vectors $\bfu$ from some prior to serve as ``true users", and simulate their query responses
assuming a logistic response model.
Regret is the difference in (true) utility of the best item under $\bfu$
and that of the item with greatest EU item given the posterior. Further
experimental details
% Details of learning 
% populational 
(e.g., on utility priors, gradient optimization configurations, data sets) can be found in Appendix~\ref{appendix:experiments}.
To reduce the effect of local EVOI maxima, we run
our continuous methods and $\qi$ with $10$ initialization
restarts (unless otherwise stated) using one of 
random, $\balanced$ or $\paolo$ initializers.

\subsection{Slate Comparisons with Complete Items}

For full-item slates, we compare to the following baselines:
\begin{description}
\item $\random$: Queries composed of randomly selected items.
\item  $\topfive$:
Exhaustive search for the best slate among the top $5$ EU items.
\item  $\greedy$: The greedy algorithm of Viappiani and Boutilier \shortcite{viappiani:nips2010}, which incrementally appends to the query slate the item that maximizes slate utility (given
items already on the slate).
\item $\qi$: The iterative algorithm from Sec.~\ref{sec:queryiteration}.
\item $\paolo$: The ``sampling initialization" heuristic of Viappiani and Boutilier \shortcite{viappiani:nips2010}, which uses the item with greatest utility for each of
$k$ randomly selected utility particles (which we find performs well on its own).
\item $\opt$: Exhaustive search (over all queries) for an optimal EVOI query.
\end{description}

\begin{figure}[t]
\includegraphics[width=\linewidth]{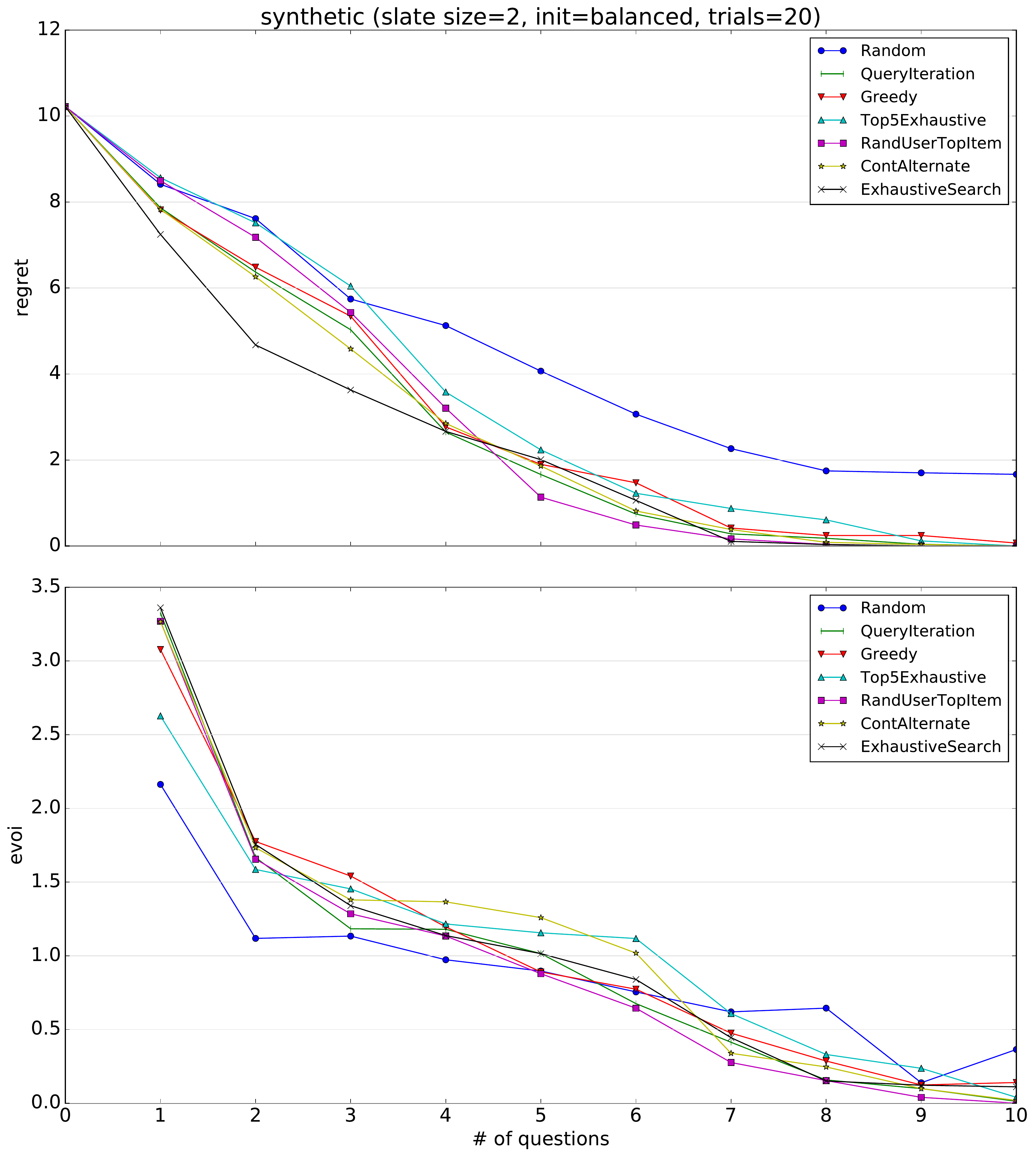}
\caption{Simulated elicitation using synthetic data.
% $\alterevoi$ followed by $\qi$ output the
% highest EVOI queries on the first question while other
% continuous methods (not shown)
% along with $\paolo$ and $\greedy$ are competitive.
\label{fig:synthetic_plot}
}
\end{figure}

{\flushleft \textbf{Synthetic}}.
We consider a synthetic dataset where we sample 
both items and utility particles with dimension $d=10$ from
$\mathcal{N}(\mathbf{0}, \mathbf{I})$. 
We run 20 trials of simulated elicitation, each sampling $5000$ items and $100$ 
utility vectors.
% CEB: REMOVE ALL OF THESE STATEMENTS... Already said so in intro above
% See Appendix~\ref{appendix:experiments} for further details.
%
Results are plotted for pairwise comparisons
in Fig.~\ref{fig:synthetic_plot} (averaged regret is an empirical estimate of ``expected loss''
as defined by Chajewska et al.~\shortcite{urszula:aaai00}). 
Of the continuous methods, only $\alterevoi$ is plotted to reduce clutter in the figure.
We first consider regret. The initial regret is
$10.45$ (no queries); and to reach regret of $1.0$, $\alterevoi$ 
takes $5$ queries (other continuous methods perform similarly) while $\qi$, $\paolo$ and $\greedy$ require
$7$ queries.
EVOI performance mirrors that of regret.
Note
that the EVOI of different methods is not generally comparable after the first query, since
different queries result in different
posteriors.
The best performing strategies for the first
query (EVOI) are $\qi$ ($3.325$), $\paolo$ ($3.269$), $\alterevoi$ ($3.266$), and $\freeevoi$ ($3.161$). $\greedy$ 
performs worst with EVOI $3.078$. Overall, these methods are competitive with $\opt$, which achieves an average EVOI of $3.361$ on the first query.
In terms of regret, $\alterevoi$ is quite competitive: achieving among the least regret across the first four queries. Because $\opt$ is myopic, it achieves the lowest regret only for the first few queries.
%

%%%% BEGIN OLD
%While the relative EVOI performance of $\alterevoi$ vs. $\greedy$ or
%$\topfive$ may seem small, of the 20 trials, $\alterevoi$
%outperforms  $\greedy$
%in $16$ trials, ties $3$ and loses $1$; and it outperforms $\topfive$
%in all $20$ trials.
%%%% END OLD
%

%
% WALL CLOCK RUNTIMES
%
\begin{table*}
\centering
\begin{tabular}{c|c|c|c|c|c|c|}
\cline{2-7}
& \multicolumn{2}{ c| }{\textbf{MovieLens}} & \multicolumn{4}{c|}{\textbf{Goodreads}} \\
\hline
\multicolumn{1}{|c|}{$(N,m,k)$} & $(1682$, $943$, $2)$ & $(1682$, $943$, $5)$ & $(2\cdot 10^5$, $100$, $2)$ & $(10^6$, $100$, $2)$ & $(10^6$, $500$, $2)$ & $(10^6$, $500$, $5)$ \\
\hline
\multicolumn{1}{|c|}{$\greedy$} & \textbf{0.01 (0.00)} & \textbf{0.04 (0.00)} & \textbf{0.14 (0.00)}	& \textbf{0.77 (0.08)} & 4.26 (0.29) & 17.62 (0.52) \\ 
\hline
\multicolumn{1}{|c|}{$\alterevoi$} & 2.56 (0.17) & 2.71 (0.09) & 2.24 (0.06) & 2.54 (0.06) & 2.96 (0.06) & 3.14 (0.06) \\ 
\hline
\multicolumn{1}{|c|}{$\freeevoi$} & 0.88 (0.04) & 0.95 (0.05) & 0.78 (0.04) & 0.85 (0.04) & \textbf{0.94 (0.04)} & \textbf{1.00 (0.04)} \\ 
\hline
\multicolumn{1}{|c|}{$\opt$} & 1+ day & N/A & N/A & N/A & N/A & N/A \\
\hline
\end{tabular}
\caption{Wall clock runtimes for EVOI algorithms. Avg. times (sec.) along with standard deviation, in parenthesis, are shown. 
%We run 10 trials with 10 queries per trial, for each algorithm and each parameter setting. The second row are parameter settings: $N$ the number of items, $m$ the numbers of particles and $k$ the slate size.
}
\label{tab:wallclock}
\end{table*}

\begin{figure}[t]
\includegraphics[width=\linewidth]{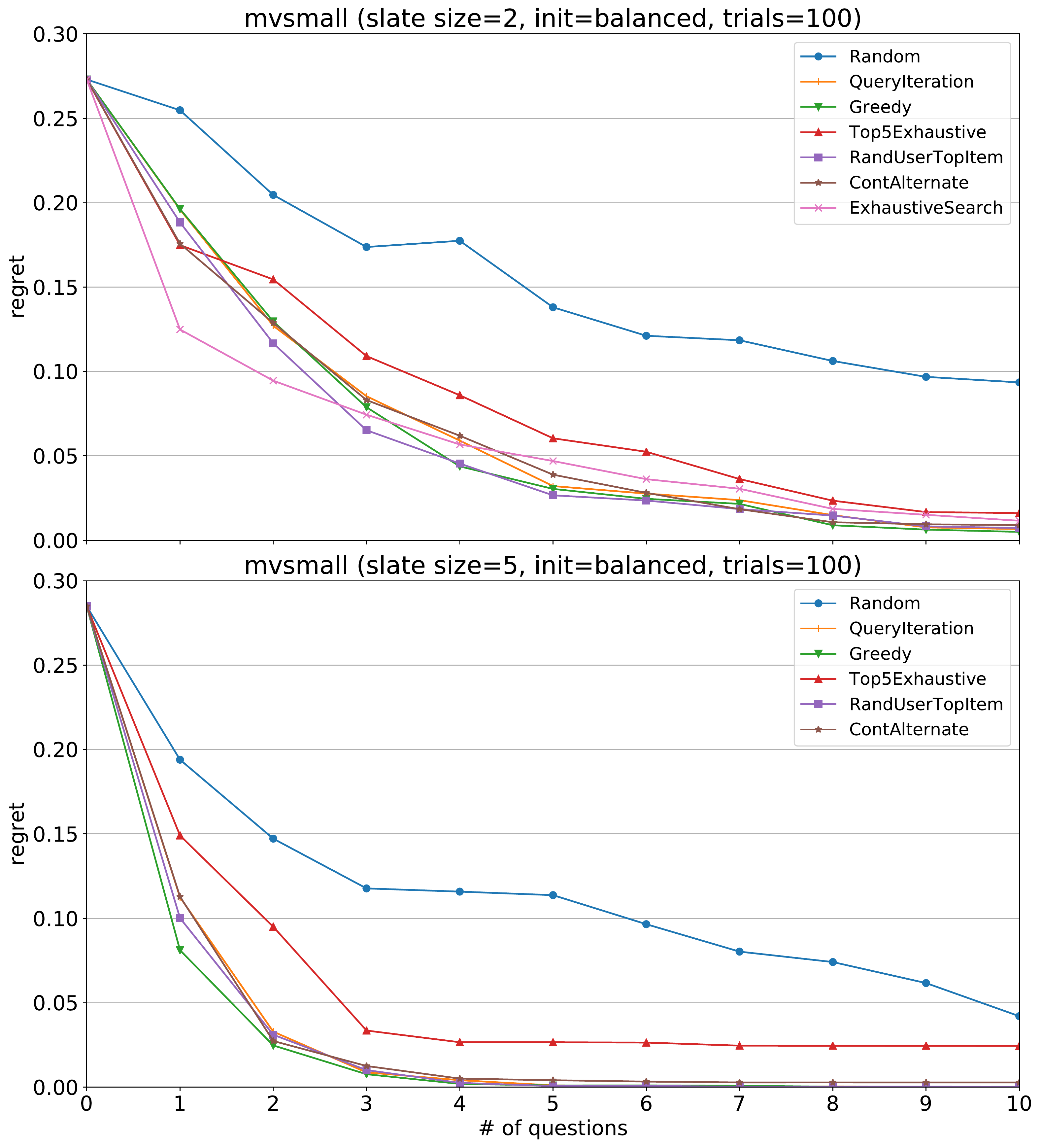}
\caption{Elicitation using MovieLens-100k.
% Methods $\drevoi$ and $\alterevoi$ are competitive with the best baselines.
\label{fig:mvsmall_slate25}}
\end{figure}

{\flushleft \textbf{MovieLens}}.
Using the MovieLens 100-K dataset,
% CEB: Not needed
% ---see  Appendix~\ref{appendix:experiments} for more details---
we train user and movie embeddings with dimension $d=10$.
Regret plots are shown in 
Fig.~\ref{fig:mvsmall_slate25} for slate sizes $k=2$ and $k=5$, averaged over $100$ elicitation trials
of $10$ queries
% for each algorithm
(using $1682$ items and random selections
of $943$ user embeddings in each trial).
Again, $\alterevoi$ is the only continuous method plotted to reduce clutter.
Regret starts at slightly below $0.3$. For pairwise comparisons, $\paolo$ takes
$5$ queries to reduce
regret to $0.03$ ($\approx 10\%$ of the original regret) while $\alterevoi$ (best continuous method), $\qi$, $\greedy$  and $\opt$ each require $6$ queries. Interestingly, while $\opt$ reduces regret the most 
% of all algorithms 
with the first two queries, it does not sustain its performance over a longer trajectory,
likely due
% to the over-optimization of
its myopic objective. 
While $\topfive$ performs best on the
first query (both w.r.t.\ regret and EVOI) it does not maintain the same regret performance (or high EVOI)
in subsequent rounds and requires $9$ queries to reach regret $0.03$. This demonstrates
one weakness of myopic EVOI and the importance of non-myopic elicitation.
For slates of size of $5$, as expected, the same top methods require about half as many queries ($3$ to $4$) to reduce regret to the same point ($0.03$).
In terms of EVOI on the first query,
for pairwise comparisons, the (in general, intractable) $\topfive$ method
performs best
with EVOI of $0.084$, while $\drevoi$ and $\alterevoi$ 
are nearly as good, with $0.084$ and $0.083$, respectively.
Next best is $\regevoi$ with $0.08$, and the remaining methods
achieve $0.0725$ or below.
For slate size $5$, $\greedy$ performs best with $0.167$,
followed by $\alterevoi$ and $\drevoi$ with $0.157$.
$\paolo$, $\qi$ and $\regevoi$ are also competitive with 
EVOI from $0.153--0.155$; meanwhile $\topfive$ performs
weaker with $0.134$ (and is worse on all
$100$ trials vs.\ all other methods).

\begin{figure}[t]
\includegraphics[width=\linewidth]{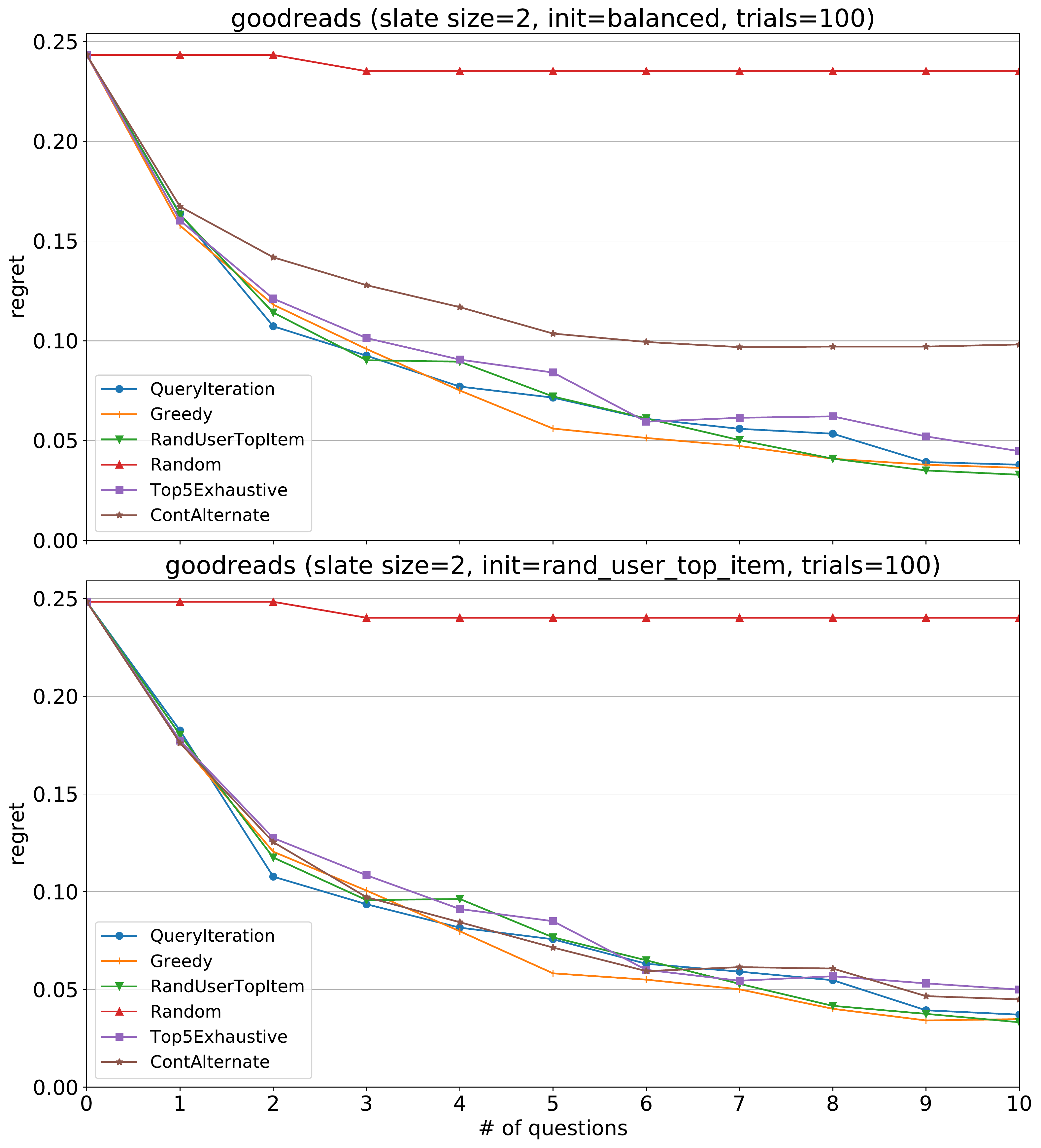}
\caption{Elicitation on Goodreads, different initialization.
% using two different initialization
% strategies shows $\balanced$ does not capture item
% space structure as well as $\paolo$.
\label{fig:goodreads_slate25}}
\end{figure}

{\flushleft \textbf{Goodreads}}. The last experiments
use the much larger Goodreads dataset.
% (see Appendix~\ref{appendix:experiments} for  more details).
We train a $d=50$ user and item embedding model.
Each trial consists of a random set of
$2\times 10^5$ items (books) and a random set of
$100$ user embeddings.
Due to the large item space, we did not run $\drevoi$ and
$\regevoi$ (it is possible to run them on subsampled item sets).
Using $\balanced$ initialization, 
the regret profile of continuous methods is not as competitive as above; e.g., in
Fig.~\ref{fig:goodreads_slate25} we see it reaches regret
$0.1$ after $5$ queries. Random performs significantly worse with this large item space. This is likely because most item pairs are not significantly different or the user embeddings tend not to vary much. 
If we instead use $\paolo$ to initialize (as $\qi$ does), we obtain competitive results w.r.t.\ both regret and EVOI.
We suspect that there could be significant structure in the
(high-dimensional) item space that volumetric initialization 
does not capture well, resulting in poor local maxima.
The choice of initialization has a similar impact on the EVOI of the first query. Our continuous methods achieve EVOI of about $0.076$ compared to $0.089$ for other methods. 

{\flushleft \textbf{Wall clock runtimes}}. We benchmark the algorithmic runtimes on a workstation with a 12-core Intel Xeon E5-1650 CPU at 3.6GHz, and 64GB of RAM. We measure the performance of $\greedy$, $\alterevoi$, and  $\freeevoi$ (with $\balanced$
initialization) since the other algorithms that are competitive in terms of EVOI are not 
computationally scalable. Results are shown in Table~\ref{tab:wallclock}. We run 10 trials with 10 queries per trial for each algorithm and under each parameter setting. Note that $\opt$ is simply not scalable beyond the smallest problem instances. We implement $\greedy$ using primarily matrix operations.
For relatively small problems $\greedy$ is fast, requiring at most $0.14$s for MovieLens and Goodreads (where the number of items, $N\le 2\cdot 10^5$). However, and as expected, scaling is an issue for $\greedy$ as it takes $4.26$s to solve for pairwise comparisons with $10^6$ items and $500$ particles/users. For a larger slate size of $5$ on Goodreads, $\greedy$ becomes even less practical, requiring $17.62$s to generate a query. Continuous methods scale better, with $\alterevoi$ taking only $3.14$s on the largest problem instance while $\freeevoi$ is even faster,
taking at most $1$s to generate a query (and is more consistent over all problem sizes)

\subsection{Slate Comparisons with Partial Items}

For partial comparison queries, we assess the quality of queries found by $\contpartial$ by comparing the EVOI of the cold-start query it finds v\`{i}s-a-v\`{i}s three natural reference algorithms: (1) random queries; (2) a natural extension of $\greedy$ for the partial setting---start with the attribute having highest utility variance across users, then greedily add attributes that result in the highest EVOI; and (3) exhaustive search for the best EVOI query.

%To evaluate $\contpartial$, 
We use the MovieLens-20M \cite{movielens:2016} dataset and represent each movie with $100$ binary attributes from the Tag Genome \cite{Vig_taggenome:2012}.
% see Appendix~\ref{appendix:experiments}.
%
%{\flushleft \textbf{EVOI Evaluation}}.
We evaluate EVOI on the first query by randomly selecting $10^5$  user embeddings as the prior, and running $\contpartial$ for 100 restarts. We initialize query embeddings to random uniform values in $[0,1]^{100}$, then run gradient ascent on Eq.~\ref{eq:partial_cont} for 100 steps, 
initializing the regularization weight $\lambda$  at $0.01$ and multiplying $\lambda$
by $1.1$ each iteration.
%

%\begin{description}
%    \item[Random:] Asks about random attributes.
%    \item[Greedy:] A natural extension of the Greedy method for full comparisons; start with the attribute that has highest variance in utility across users, then greedily add the attribute that gives the highest EVOI query.
%    \item[Best Possible:] Best possible partial slate query, computed by exhaustive search for slate sizes where this is computationally feasible. 
%\end{description}
%
\begin{figure}[t]
\begin{tikzpicture}
\begin{axis}[width=9cm, height=5cm,
    xlabel={Query Slate Size},
    ylabel={EVOI},
    xmin=1, xmax=6.5,
    ymin=0, ymax=0.93,
    xtick={2,3,4,5,6},
    ytick={0.2, 0.4, 0.6, 0.8},
    legend pos=north west,
    ymajorgrids=true,
    grid style=dashed,
]

%\errorband[black, %opacity=0.2]{partial_random_1attr.dat}{size}{low%er}{upper}
\errorband{partial_random_1attr.dat}{size}{evoi}{lower}{upper}{black}{square}
\addlegendentry{\tiny{$\random$}}

%\addplot color=black,
%     mark=square] table [
%     size index=0,
%     evoi index=1];% {partial_random_1attr.dat};
   % \addlegendentry{Random}

% \addplot[
%     color=black,
%     mark=square,
%     ]
%     coordinates {
%     (2, 0.0503)(3, 0.1217)(4,0.1787)(5,0.2239)(6,0.2519)
%     };
%     \addlegendentry{Random}

\addplot[
    color=green,
    mark=square,
    ]
    coordinates {
    (2, 0.1651)(3, 0.3224)(4,0.3895)(5,0.4472)(6,0.507)
    };
    \addlegendentry{\tiny{$\greedy$}}

\addplot[
    color=blue,
    mark=*,
    ]
    coordinates {
    (2, 0.2831)(3,0.4581)(4,0.5480)
    };
    \addlegendentry{\tiny{$\opt$}}

\errorband{cont_partial_1attr.dat}{size}{evoi}{lower}{upper}{red}{triangle}
    \addlegendentry{\tiny{$\contpartial$}}
    
% \addplot[
%     color=red,
%     mark=triangle,
%     style=dashed,
%     ]
%     coordinates {
%     (2, 0.4524)(3, 0.6583)(4, 0.7294)(5, 0.7937)(6,0.8778)
%     };
%     \addlegendentry{5-attr Continuous Indicator}

% \addplot[
%     color=purple,
%     mark=square,
%     style=dashed,
%     ]
%     coordinates {
%     (2, 0.6251)(3, 0.9031)(4,1.083)(5,1.2044)(6,1.278)
%     };
%     \addlegendentry{Full Greedy}

\end{axis}
\end{tikzpicture}
\caption{Single-attribute EVOI of queries found by $\contpartial$ vs baselines. For methods with randomness we show the mean and interquartile range over 100 trials. Computational constraints only allow us to compute Best Possible for $k\le 4$.}
\label{fig:partial}
\end{figure}
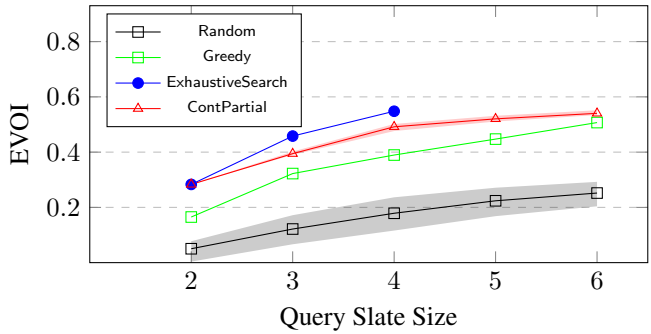

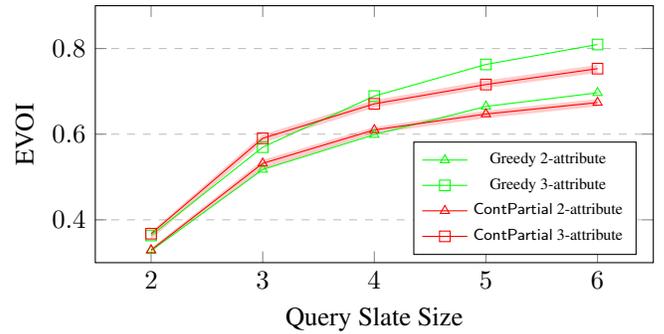
\begin{figure}[t]
\begin{tikzpicture}
\begin{axis}[width=9cm, height=5cm,
    xlabel={Query Slate Size},
    ylabel={EVOI},
    xmin=1.5, xmax=6.5,
    ymin=0.3, ymax=0.9,
    xtick={2,3,4,5,6},
    ytick={0.4, 0.6, 0.8},
    legend pos=south east,
    ymajorgrids=true,
    grid style=dashed,
]

%\errorband[black, %opacity=0.2]{partial_random_1attr.dat}{size}{low%er}{upper}
%\errorband{partial_random_2attr.dat}{size}{evoi}{lower}{upper}{black}{square}
%\addlegendentry{\tiny{Random 2-attribute}}

\addplot[
    color=green,
    mark=triangle,
    ]
    coordinates {
    (2, 0.3278)(3, 0.5175)(4,0.5989)(5,0.6648)(6,0.6966)
    };
    \addlegendentry{\tiny{Greedy 2-attribute}}

\addplot[
    color=green,
    mark=square,
    ]
    coordinates {
    (2, 0.3634)(3, 0.5696)(4,0.6890)(5,0.7628)(6,0.8094)
    };
    \addlegendentry{\tiny{Greedy 3-attribute}}

\errorband{cont_partial_2attr.dat}{size}{evoi}{lower}{upper}{red}{triangle}
    \addlegendentry{\tiny{$\contpartial$ 2-attribute}}
    
\errorband{cont_partial_3attr.dat}{size}{evoi}{lower}{upper}{red}{square}
    \addlegendentry{\tiny{$\contpartial$ 3-attribute}}

\end{axis}
\end{tikzpicture}
\caption{Multi-attribute EVOI of queries found by $\contpartial$ vs our greedy baseline.}
\label{fig:partial_multiattribute}
\end{figure}

As Figs.~\ref{fig:partial} and \ref{fig:partial_multiattribute} show, $\contpartial$ outperforms greedy and comes close to finding the best query for smaller slates and single attributes. With larger slates and multiple attributes, greedy performs better.
Appendix~\ref{appendix:experiments} shows an example elicitation tree and recommendations found by $\contpartial$.

\section{Conclusion and Future Work}
\label{sec:conclude}

We have developed a continuous relaxation for EVOI optimization for
Bayesian preference elicitation in RSs
that allows scalable, flexible gradient-based approaches.
% towards preference elicitation which allows for more practical and flexible interactive % recommendations. 
We also leveraged this approach to develop an EVOI-based algorithm
for partial comparison queries.
% which we propose to mitigate both the cognitive burden and irrelevance of full comparisons in many realistic scenarios.
Our methods readily handle domains with large item spaces, continuous
item attributes,
% number of items, partial comparison queries on a smaller subset of a much wider set of attributes while also being readily 
and can be adapted to other differentiable metrics.
% over the recommendation space.
They can also leverage modern ML
frameworks to exploit various algorithmic, software
and hardware efficiencies. Our experiments show that continuous EVOI achieves state-of-the-art results in some domains.
% and is particularly useful for partial comparisons.

There are various avenues for future work. Further exploration of
different forms of partial comparisons is of interest,
including the use of \emph{latent} or high-level conceptual features while using continuous elicitation methods to generate informative queries
from a much larger, perhaps continuous, query space.
Methods that
incorporate user knowledge, natural language modeling and
visual features, together with explicit or latent attributes
during elicitation would be of great value. Finally,
evaluating recommendations using traditional ranking metrics and conducting
user studies will play a key role in making elicitation more user-friendly.

%Elicitation over images / language in an embedding space.
%Evaluate on traditional ranking metrics.
%Other response models for partial queries.

% \bibliography{long,standard}
% \bibliographystyle{aaai}

\clearpage
\appendix

\section{Appendix: Supplementary Material}

\subsection{Proof Sketch of Theorem~\ref{thm:viappiani}}
\begin{proof}[Proof Sketch]
Without loss of generality, assume $k=2$. Under hardmax we have:
\begin{align}
  \peu(q) &= \int {\bfy^*_1}^T 
    \bfu \underbrace{P(r=\bfx_1 | \bfu ; q)}_{=:w_1} P(\bfu) d\bfu  + \nonumber \\
    & 
    \int {\bfy^*_2}^T 
    \bfu \underbrace{P(r=\bfx_2 | \bfu ; q)}_{=:w_2} P(\bfu) d\bfu \nonumber \\
  &= \int \left[ w_1{\bfy^*_1}^T \bfu +
                 w_2{\bfy^*_2}^T \bfu \right]
          P(\bfu) d\bfu \label{eq:evoi_iter_max}
\end{align}
The best weighting to maximize Eq.~\ref{eq:evoi_iter_max} is to set
$w_1=1$ and $w_2=0$ when $u({\bfy^*_1}^T ; \bfu) = {\bfy^*_1}^T \bfu \ge 
{\bfy^*_2}^T  \bfu = u({\bfy^*_2}^T ; \bfu)$. Likewise set $w_2=1$ and
$w_1=0$ when ${\bfy^*_1}^T \bfu < {\bfy^*_2}^T\bfu$. These two
conditions are exactly what $w_1=P(r=\bfx_1 | \bfu ; q)$ and 
$w_2=P(r=\bfx_2 | \bfu ; q)$ achieves under hardmax. Extension of this
argument to $k$-wise slate response is similar. Under softmax, 
there is a slight dip (depending on temperature
and item space) in the probability from $1$.
\end{proof}

\subsection{Description of Continuous EVOI Algorithms}
\label{appendix:algos}
The optimization objective in Alg.~\ref{alg:contbase} is non-convex but can be formulated as training a supervised
learning problem where the $\bfu_j$'s are examples with a dummy label,
and passed through a neural network with softmax activations multiplied by the output of $\bfY^T \bfU$. 
Using standardized machine learning 
frameworks such as TensorFlow or PyTorch one can apply gradient-based
optimization algorithms.

The baseline algorithm we presented in Alg.~\ref{alg:contbase}
does not use any structure (or knowledge) of the feasible 
item space until the final deep retrieval operation. This can cause
the optimization to potentially reach regions of $\R^d$ that are 
far from any realizable items. There are a few ways to
impose item structure for query slate $\bfX$ and/or 
the rec. slate $\bfY$.

One can
incorporate feasibility constraints as regularization terms in the
optimization. For example, we can add a smooth regularizer 
$\lambda \sum_{i=1}^k \min_{\bfz \in \calX} \| \bfx_i - \bfz \|^2_2$ which
ensures closeness to feasible items. We call this algorithm $\regevoi$
(see Alg. \ref{alg:regevoi}).
However, this regularization term can magnify local optimality issues,
since optimization variables $\bfx_i$'s prefer to stay around existing
feasible items). From a running time perspective, the min function has
to enumerate over a potentially large item space. 
%We discuss ways of reducing the item space below.

Instead of regularizing, we can also directly optimize the deep 
retrieved items. That is, we can maximize the PEU resulting 
from a deep retrieval
operation, $\bfy_1 \ldots \bfy_k = \DR(\bfx_1, \ldots, \bfx_k)$ for each of the
possible $k$ slate query responses, multiplied by the softmax response
model. This enables more accurate representation of the dot products
in the EVOI objective. 
Note that this operation is a hard-max over all items,
which is still (mostly) differentiable (we can also implement
a soft version, e.g. by using $\ell_p$ norm for sufficiently large
$p$).\footnote{Likewise for gradient optimization we can also use hardmax probabilities in place of the softmax probabilities of the response model.}

Because a corollary of Thm.~\ref{thm:viappiani} is that
there exists an optimal feasible query slate (for hardmax; for softmax
a close to optimal slate exists), we can in practice remove any norm
constraints on $\bfx_i$---since during optimization, $\bfX$ variables 
will naturally gravitate towards feasible items as it can achieve 
higher objective values (for reasonable softmax temperatures). By directly
optimizing the deep retrieved PEU, the output ``loss'' is more reflective 
of the actual PEU, the only approximation error comes from the free variables
$\bfx_i$ appearing in the softmax. This algorithm is called
$\drevoi$. However, as with regularization, an 
explicit max enumeration over feasible items is required in the 
deep retrieval objective.

A final variant of continuous EVOI tries to account for item structure
without explicit enumeration of items into the optimization. It is
similar to $\freeevoi$ except we iteratively optimize $\bfx_i$ (subject
to norm bounds), then deep retrieve with $\druniq$ 
(making sure duplicate items
are replaced by the distinct items with the next highest expected
utilities)
for $\bfy_i^*$, which we call
$\alterevoi$ since it alternatively optimizes the query slate ($\bfx_i$'s)
and the deep retrieved recommendation slate ($\bfy_i$'s).
%
% FREE EVOI
%
\begin{algorithm}[t]
 \caption{$\freeevoi$, Continuous Free EVOI}
 \label{alg:freeevoi}
 \begin{algorithmic}[1]
   \STATE Initialize $\bfX$ (e.g. with random, $\balanced$ or $\paolo$)
   \STATE Similar to Alg.~\ref{alg:contbase} but using only one set of variables $\bfX$ to also represent recommendation slate $\bfY$.
   \STATE One can optimize $\bfX$, subject to norm bounds (max norm, unit norm, etc.).
 \end{algorithmic}
\end{algorithm}
%
% REGULARIZED EVOI
%
\begin{algorithm}[th]
 \caption{$\regevoi$, Continuous Regularized EVOI}
 \label{alg:regevoi}
 \begin{algorithmic}[1]
   \STATE Same as $\freeevoi$ but with an added penalty term in the objective: $\lambda \sum_{i=1}^k \min_{\bfz \in \calX} \| \bfx_i - \bfz \|^2$ 
 \end{algorithmic}
\end{algorithm}
%
% ALTERNATING EVOI
%
\begin{algorithm}[th]
 \caption{$\alterevoi$, Continuous Alternating EVOI}
 \label{alg:alterevoi}
 \begin{algorithmic}[1]
   \STATE Initialize $\bfX$ and $\bfY$    
   \STATE $\bfp = \mathrm{softmax}(\bfU \bfX)$
   \FOR{$i=1..$ {\tt max\_iterations}}
   \STATE $\bfX \leftarrow \max_{\|\bfX\|\le B} \1^T_m \bfU\bfY \circ \mathrm{softmax}(\bfU\bfX)\1_k $ using gradient methods.
   \STATE $\bfY \leftarrow \DR(\bfX)$, i.e., each item in the slate for $\bfY$ is a deep retrieval with respect to query slate $\bfX$.
   \ENDFOR
   \RETURN $\druniq(\bfX, \bfU)$
 \end{algorithmic}
\end{algorithm}
%
% DEEP RETRIEVAL EVOI
%
\begin{algorithm}[t]
 \caption{Deep Retrieval EVOI}
 \label{alg:drevoi}
 \begin{algorithmic}[1]
   \STATE $\bfX \leftarrow \max_\bfX \1^T_m \mathtt{max}(\mathrm{softmax}(\bfU\bfX)^T \bfU \calX^T)$ using gradient methods, where $\mathtt{max}$ returns the maximum value for each row vector of the input matrix.
   \RETURN $\druniq(\bfX, \bfU)$
 \end{algorithmic}
\end{algorithm}

\subsection{Experiments}
\label{appendix:experiments}
We use the standard TensorFlow implementation of Adam
\cite{kingma-ba:iclr2015} to implement our continuous
algorithms. 

{\flushleft \textbf{Synthetic Dataset}}.
We used a learning rate of $0.0005$. For the softmax temperature during optimization, we used $0.02$ while the softmax temperature we used for evaluation is $0.1$.

{\flushleft \textbf{MovieLens-100k Dataset}}.
We trained $d=10$ user and movie embeddings via Probabilistic
Matrix Factorization \cite{prob-CF:08}, as implemented by 
\cite{pmf_code:github2016} on the MovieLens-100k dataset 
\cite{movielens:2016}.
We used a learning rate of $0.001$ with
a softmax temperature of $0.03$ during optimization and a softmax temperature of $0.01$ during evaluation. 

{\flushleft \textbf{Goodreads}}
This is a much larger dataset consisting of user interactions with
the Goodreads website~\cite{wan2018goodreads}. There are
about $2\times 10^6$ items and about $10^6$ users. We used the user ratings
to learn a $d=50$ user and item embedding model using the commonly used alternating 
least-squares method of \cite{Hu:2008} with the conjugate-gradient optimizations of \cite{Takacs:2011}.
For continuous methods,
we used a learning rate of $0.0005$ and softmax temperature of $0.02$ for
optimization and a softmax temperature of $0.1$ for evaluation.

{\flushleft \textbf{Partial Comparisons: MovieLens-20M}}
\begin{figure*}[th]
\includegraphics[width=\textwidth]{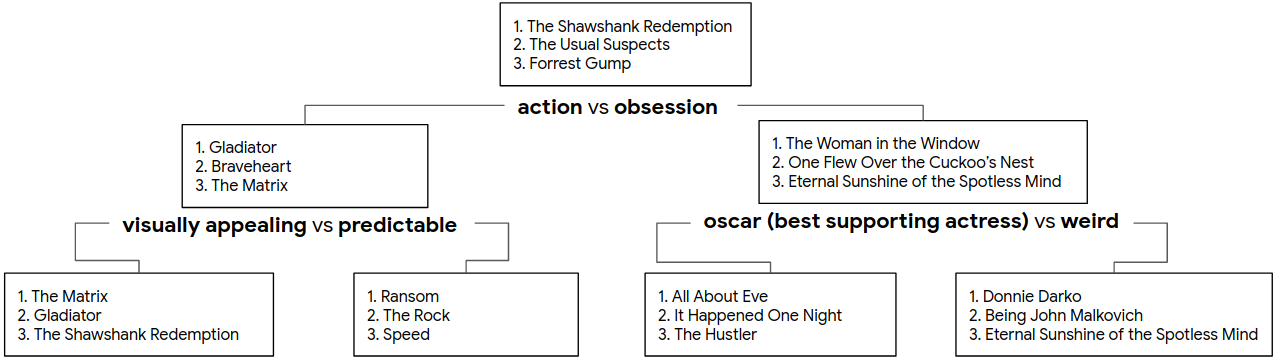}
\caption{Elicitation tree found by $\contpartial$ for partial pairwise single-attribute queries. The top 3 movies are displayed at each point, starting with cold-start users at the top. Note how tags like ``predictable", while highly informative, are probably not how users who like those movies would describe their preferences; handling \emph{personalized language}, how different users use different words to describe their preferences, is a promising direction of future work.\label{fig:mv_tree}}
\end{figure*}
We use MovieLens-20M \cite{movielens:2016} 
a dataset of $2\times 10^6$ movie ratings, and the Tag Genome \cite{Vig_taggenome:2012}, which annotates each movie with relevance scores between 0 and 1 for 1000 binary attributes (or \emph{tags}) such as ``action", ``weird", ``based on a book".
We took the 100 most common tags from the Tag Genome (measured by sum of relevance scores across movies), and filtered out movies with less than 10 ratings or with so few tags that their relevance scores sum to less than 10, with 10307 movies remaining. We then represented movies as 100-dimensional vectors of their relevance scores for each tag.
We trained user embeddings to predict movies users liked by maximizing for each user $\bfu$:
$$
\max_{\bfu \in \R^{100}} 
\sum_{\bfp \in \textrm{Pos}, \mathbf{n} \in \textrm{Neg}} \frac{e^{\bfu^T\bfp}}{(e^{\bfu^T\bfp} + e^{\bfu^T\mathbf{n}})}
$$
where $\textrm{Pos}$ are movies the user rated at least $4$ out of $5$, and $\textrm{Neg}$ are randomly sampled from movies other users rated.

The user embeddings were trained with Adam, learning rate $10^{-3}$, batch size = $10000$, L2 regularization parameter $\alpha=10^{-6}$. 

To prevent overfitting we withhold $20\%$ of each user's ratings as validation set, and ended training when validation binary accuracy plateaued. Final binary accuracy on the validation set was $73\%$, while the popularity baseline (ranking movies by their number of ratings) is $54\%$.

% [optional, why this instead of matrix factorization?] 
%Embeddings learned this way lead to considerably more diversity than those learned using matrix factorization methods such as ALS (TODO add ref), which are dominated by popular movies. The dot-product utilities also more plausibly reflect true user utilities since they are learned by predicting user's movie watches assuming the Placcett-Luce model of user choice (TODO add ref).

We ran gradient descent on Eq. \ref{eq:partial_cont}
 and multiplied the regularization weight $\lambda$ by $1.1$ each iteration, starting from $0.01$. The softmax temperature for both optimization and evaluation was $0.1$. All hyperparameters were tuned on a separate set of $10000$ users.
 
To give an illustration for how elicitation using our method would work in practice, 
Fig.~\ref{fig:mv_tree} shows the elicitation queries and recommendations found by $\contpartial$,
for the single-attribute pairwise comparison case.
\end{document}